\documentclass[10pt,twocolumn,letterpaper]{article}

\usepackage{pifont}
\usepackage{comment}
\usepackage{multirow}
\usepackage[pagenumbers]{cvpr} 

\usepackage[dvipsnames]{xcolor}

\definecolor{cvprblue}{rgb}{0.21,0.49,0.74}
\usepackage[pagebackref,breaklinks,colorlinks,citecolor=cvprblue]{hyperref}
\usepackage[marginal]{footmisc}

\def\paperID{3629} 
\def\confName{CVPR}
\def\confYear{2024}
\newcommand{\wlink}[1]{\textcolor{magenta}{{#1}}}
\title{PICTURE: PhotorealistIC virtual Try-on from UnconstRained dEsigns} 

\author{
    Shuliang Ning\textsuperscript{1,2,3}   \qquad
    Duomin Wang\textsuperscript{3} \qquad
    Yipeng Qin\textsuperscript{4} \qquad  
    Zirong Jin\textsuperscript{2} \qquad \\
    Baoyuan Wang\textsuperscript{3} \qquad
    Xiaoguang Han\textsuperscript{2,1,\footnotemark[1]} \\
\textsuperscript{1} FNii, CUHKSZ \qquad
\textsuperscript{2} SSE, CUHKSZ \qquad
\textsuperscript{3} Xiaobing.AI \qquad
\textsuperscript{4} Cardiff University \qquad
}

\begin{document}
\twocolumn[{
\renewcommand\twocolumn[1][]{#1}
\maketitle
\begin{center}
    \captionsetup{type=figure} 
    \includegraphics[width=.93\textwidth]{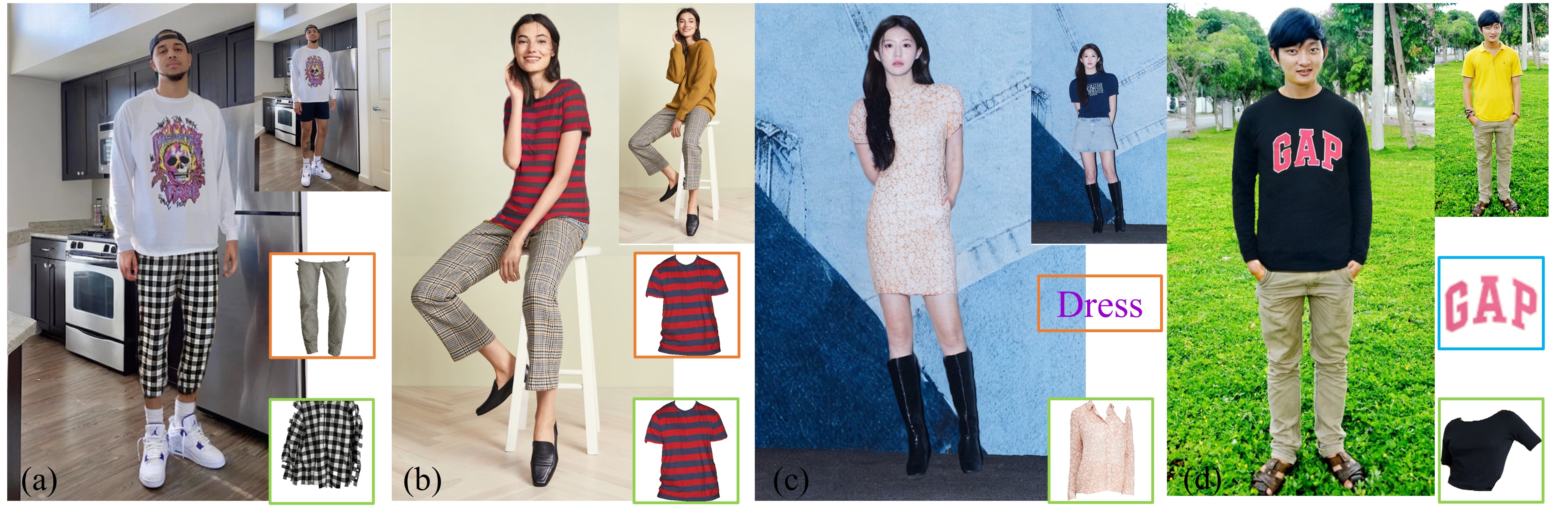}
    \vspace{-2mm}
    \caption{Examples of virtual try-on manipulations using ucVTON on \textbf{in-the-wild} real images. Orange box: style control; Green box: texture control; Blue box: design elements control.  }
    \label{fig:teaser}
    \vspace{-2mm}
\end{center}
}]
\footnotetext[1]{Corresponding author:
\wlink{hanxiaoguang@cuhk.edu.cn}.}
\begin{abstract}
\vspace{-2mm}
In this paper, we propose a novel virtual try-on from unconstrained designs (ucVTON) task to enable photorealistic synthesis of personalized composite clothing on input human images.  Unlike prior arts constrained by specific input types, our method allows flexible specification of style (text or image) and texture (full garment, cropped sections, or texture patches) conditions. To address the entanglement challenge when using full garment images as conditions, we develop a two-stage pipeline with explicit disentanglement of style and texture. In the first stage, we generate a human parsing map reflecting the desired style conditioned on the input. In the second stage, we composite textures onto the parsing map areas based on the texture input. 
To represent complex and non-stationary textures that have never been achieved in previous fashion editing works, we first propose extracting hierarchical and balanced CLIP features and applying position encoding in VTON.
Experiments demonstrate superior synthesis quality and personalization enabled by our method. The flexible control over style and texture mixing brings virtual try-on to a new level of user experience for online shopping and fashion design.

\end{abstract}
\vspace{-0.6cm}    
\section{Introduction}
\label{sec:intro}
Virtual try-on (VTON) systems have become indispensible in the era of online clothing shopping and are an active area of research in computer vision. 
As online shopping grows increasingly ubiquitous in modern digital lifestyles, VTON addresses a key limitation: the inability to physically try on clothes prior to purchase. 
By enabling users to visualize garments on their photos or avatars, VTON systems aim to provide crucial visual information about fit and appearance. 

Traditional VTON~\cite{yu2019vtnfp,dong2019towards,yang2020towards,choi2021viton,ge2021parser,bai2022single,morelli2022dress,he2022style,lee2022high, morelli2023ladi, gou2023taming} focuses on in-shop scenarios which create photorealistic visualization by generating images of individuals wearing existing retail garments. 
This is constrained by input person and clothing images so that the person's identity and appearance are retained and the generated clothing matches the input. 
However, this constrained generation is also limited, as real-world online shopping behaviors often involve a bit of design, with a desire to mix-and-match elements from different garments. 
That is, users may wish to visualize personalized composite clothing items, combining preferred style and texture aspects from separate pieces.
Enabling such controllable synthesis of new clothing on virtual try-on remains an open challenge. 
FashionTex~\cite{lin2023fashiontex} pioneered this strand by using a disentangled (style, texture) representation to control the VTON synthesis.
However, despite its success, FashionTex is still constrained by the type of its inputs: 
i) the style condition is limited to text prompts, which are less accessible and inclusive to users with dyslexia or other language barriers;
ii) the texture input is limited to small image patches, which cannot characterize complex fabrics or patterns.
Both constraints reduce its real-world applicability.

In this work, we fill this gap by proposing a new task called VTON from unconstrained designs (ucVTON), which greatly relaxes the constraints mentioned above, enabling users to specify style via images, and texture via full-garment images or swatches.
Specifically, we expand the allowable input types for increased flexibility: the style condition can be a text prompt or an example garment image; the texture condition can be a full garment, a cropped section of a garment, or an image patch. 
This allows representing complex fabric textures across different scales and spatial distributions, bringing fine-grained control of VTON to a new level of synthesis quality and user personalization.
However, a significant challenge emerges when enabling style and texture inputs to be full garment images: both inputs contain entangled style and texture features, and the network can struggle to disentangle the irrelevant style from the texture input and irrelevant texture from the style input for proper re-composition.

Addressing this challenge, we propose a novel two-stage style and texture disentanglement pipeline based on Stable Diffusion~\cite{rombach2022high}, where the first stage explicitly learns to generate clothing styles in the format of a human parsing map conditioned on a given style input; the second stage then adds textures to the garment areas of the parsing map conditioned on the given texture input.
In addition, instead of naively replacing the text CLIP features used in previous methods with image CLIP features, we propose a novel approach based on our observation that the final-block CLIP features are rich in semantics but lack low-level texture features. Specifically, we propose to learn photorealistic texture features from the texture input through a hierarchical and balanced combination of CLIP features from multiple blocks of its Vision Transformer.
Finally, we utilize position encoding~\cite{vaswani2017attention} to represent the spatial distributions and scales of non-stationary input textures. 
This significantly expands the space of possible texture inputs that can be handled, which has never been achieved by previous methods.
Extensive experimental results demonstrate the effectiveness of our method. 
Our contributions include:
\begin{itemize}
\item  We introduce the novel task of virtual try-on from unconstrained designs (ucVTON), which significantly advances the state-of-the-art by enabling photorealistic VTON from diverse style (text prompts or garment images) and texture inputs (full garments, cropped sections of a garment, or image patches). 
\item We propose a two-stage disentanglement pipeline that explicitly separates style generation and texture composition for controllable ucVTON from entangled inputs.
\item We first propose a new hierarchical CLIP feature extraction and position encoding method to represent photorealistic, non-stationary textures. This significantly expands the diversity and complexity of synthesized textures that has never been achieved by previous methods.
\item Extensive experiments demonstrate that our method has brought fine-grained control of VTON to a new level of synthesis quality and user personalization.
\end{itemize}
\vspace{-0.2cm}

\section{Related Work}
\label{sec:related}
\noindent\textbf{Image-based Fashion Editing.}
Given a human image and some editing conditions, such as reference cloth images~\cite{morelli2023ladi,gou2023taming,chen2020tailorgan,yang2023paint,huang2023composer} or text descriptions of styles~\cite{jiang2022text2human,baldrati2023multimodal,patashnik2021styleclip,xia2021tedigan,alaluf2022hyperstyle}, image-based fashion editing aims to generate a target image that satisfies the given conditions while preserving the rest (\eg, identity,  skin) of the original human image.
Examples of existing solutions include TextureReformer~\cite{wang2022texture}, which employs a multi-view, multi-stage synthesis procedure to perform interactive texture transfer under user-specified guidance;
Text2Human~\cite{jiang2022text2human}, which implements a two-step process using text descriptions to synthesize human images from a given pose, focusing on clothing shapes and textures;
FashionTex~\cite{lin2023fashiontex}, which develops a fashion editing module that harnesses both text and texture inputs for multi-level fashion editing in full-body portraits.

In this work, we follow the multimodal setup of FashionTex~\cite{lin2023fashiontex} and take both text and example images as inputs. 
However, unlike~\cite{lin2023fashiontex} which assumes the input texture to be a small image patch (64$\times$64) with stationary textures, we have made significant improvements to the model to enable it to handle unconstrained high-resolution garment images with arbitrary textures, shapes, positions, and scales as input. This allows for higher quality, flexibility and control over the generated results compared to prior arts. 

\vspace{2mm}
\noindent\textbf{Diffusion Models.}
Diffusion models~\cite{song2020denoising,ho2020denoising,rombach2022high} are a major type of deep generative models that exhibit robustness and superior proficiency in handling diverse data modalities~\cite{zhao2023uni}. 
Inspired by \cite{rombach2022high,mou2023t2i,zhang2023adding}, numerous conditional image generation works~\cite{xiu2023econ,kim2023dense,goel2023pair,wu2023harnessing,xue2023freestyle,ren2023multiscale,saini2023chop} have incorporated diffusion models into their methods. 
For example, SGDiff~\cite{sun2023sgdiff} introduces a style-guided diffusion model that combines text and style images to facilitate creative image synthesis; 
Paint-by-Example~\cite{yang2023paint} leverages self-supervised training to disentangle and re-organize the source image and the exemplar for more precise control in exemplar-guided image inpainting; Multimodal Garment Designer~\cite{baldrati2023multimodal} employs a latent diffusion model and multimodal conditions such as text, pose map, and garment sketches for fashion editing. 
Their impressive results demonstrate the power of diffusion models as a solution for high-quality, controllable image generation and editing tasks.



\section{Background and Problem Definition}
\label{sec:background}

\begin{table}
  \centering
\setlength\tabcolsep{5pt}
  \begin{tabular}{|c|c|c|c|c|c|}
    \hline 
    Method &  $C_s^{text}$ & $C_s^{img}$ &  $C_t^{patch}$ & $C_t^{img}$  \\
    \hline 
    Texture Reformer\cite{wang2022texture} & \ding{53} & \ding{53} & \ding{51}& \ding{51} \\
    PIDM\cite{bhunia2023person}  & \ding{53} & \ding{53} & \ding{53}& \ding{51} \\
    FashionTex\cite{lin2023fashiontex} & \ding{51} & \ding{53} & \ding{51}& \ding{53} \\
    Text2Human\cite{jiang2022text2human} & \ding{51} & \ding{53} & \ding{53}& \ding{53} \\
    DCI-VTON\cite{gou2023taming} & \ding{53} & \ding{51} & \ding{51}& \ding{53} \\
    \hline 
    {\bf Ours} & \ding{51} &  \ding{51}& \ding{51}& \ding{51} \\
    \hline 
  \end{tabular}
  \caption{Comparison with SOTA methods on input types allowed.}
    \vspace{-0.3cm}
  \label{tab: conditional flexibility }
\end{table}

Traditionally, virtual try-on can be defined as:
\begin{equation}
    \hat{I} = \mathrm{VTON}(I, G) 
\end{equation}
where $I$ denotes the input image of a person, $G$ denotes the garment image to be virtually worn, $\hat{I}$ is the output image showing the person wearing garment $G$.
Although straightforward, this definition is limited because it requires a {\it complete} garment image $G$ with a fixed combination of clothing style and texture. 
This contradicts people's imagination ability to envision wearing a particular clothing style with different textures.
Addressing this limitation, FashionTex~\cite{lin2023fashiontex} pioneered in disentangling $G$ into style condition $C_s^{text}$ and texture condition $C_t^{patch}$ and have:
\begin{align}
\begin{split}
    &\hat{I} = \mathrm{VTON}(I, \{C_s^{text}, C_t^{patch}\}) 
\end{split}
\end{align}
where $C_s^{text}$ is a text prompt describing the garment style and $C_t^{patch}$ is a small image patch describing the local texture of the garment.
Despite its success, FashionTex is constrained by the type of inputs in that: i) the style condition is limited to text prompts $C_s^{text}$, which are less accessible and inclusive to users
with dyslexia or other language barriers; 
ii) the texture input is limited to small image patches $C_t^{patch}$, which cannot characterize complex fabrics or patterns. 
These constraints reduce its real-world applicability.
For practical virtual try-on, users need more flexible condition types - the ability to specify style via images
, and texture via full-garment images or swatches.
To fill this gap, in this work, we propose virtual try-on from {\bf unconstrained designs} (ucVTON):
\begin{equation}
    \hat{I} = \mathrm{ucVTON}(I, \{C_s^{text/img}, C_t^{img/patch}\}) 
\end{equation}
In this way, we expand the allowable condition types for style and texture as follows: the style condition can be either a text prompt or an example garment image, providing more flexibility in specifying the desired design. The texture condition can be a full-garment image, a cropped section of a garment, or an image patch, enabling the representation of complex textures at varied scales and distributions. This enhanced flexibility empowers users with finer-grained control over both style and texture.
Please see Table~\ref{tab: conditional flexibility } for a comparison of our ucVTON and SOTA methods.

\section{Method}
\label{sec:method}

\begin{figure*}[t]
	\centering
	\includegraphics[width=0.85\linewidth]{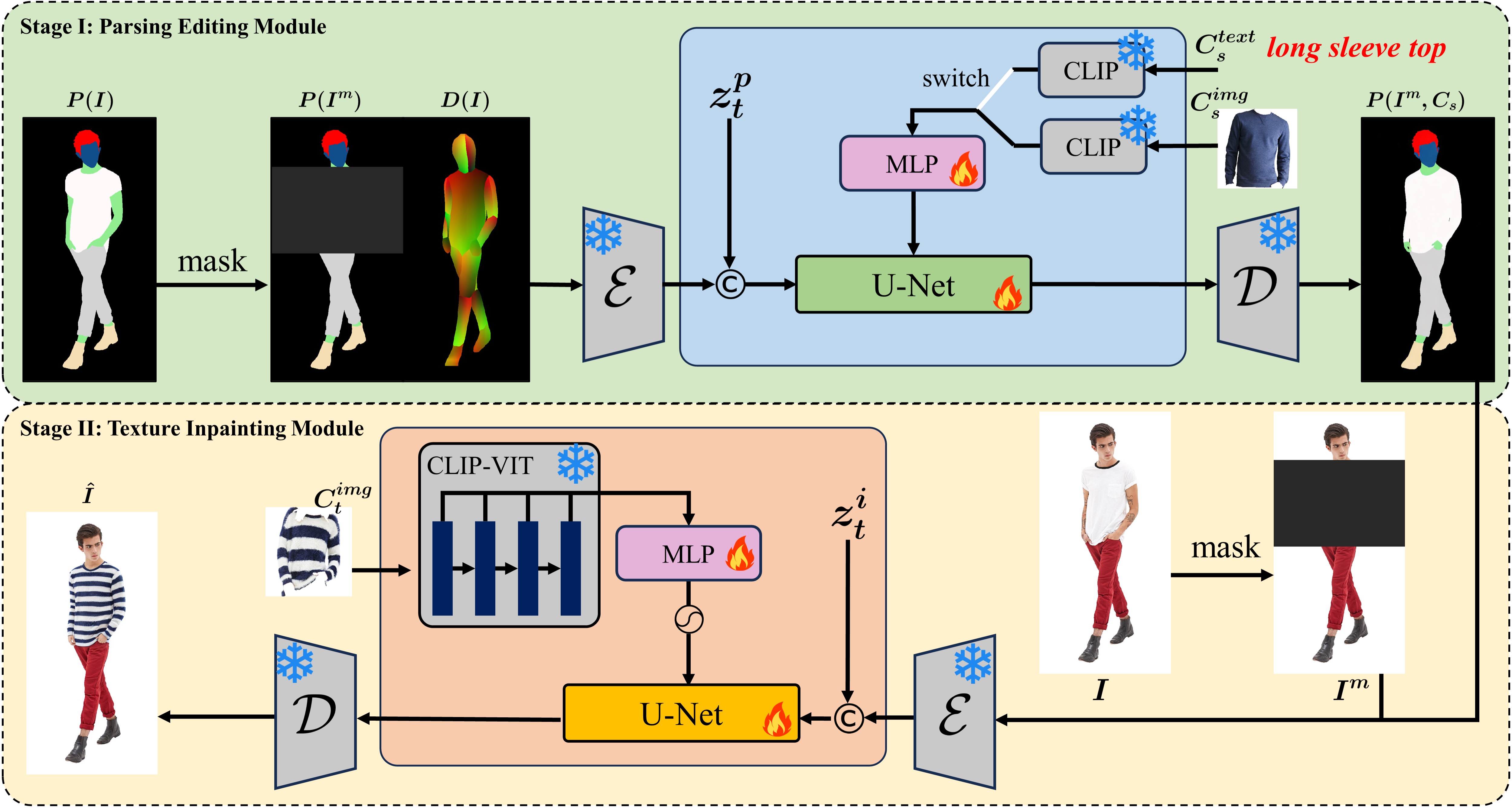}
	\caption{Overall pipeline. 
 {\bf Stage I}: Given a clothing-agnostic parsing image $P(I^m)$ and its corresponding densepose image $D(I)$, a style condition $C_{s}$ (text or image), this stage generates a parsing map $P(I_m, C_s)$ edited according to $C_{s}$.  
 {\bf Stage II}: Given a clothing-agnostic human image $I^m$, the parsing map $P(I_m, C_s)$ generated in Stage I, a garment texture condition $C_{t}$ (image or patch), this stage generates the final human image $\hat{I}$ with its style specified by $C_s$ and texture specified by $C_{t}$. 
 Note that we only used $C_{t}^{img}$ during training and the learned disentanglement allows for using $C_{t}^{patch}$ during inference.
 Flame symbol: ``trainable''; Snowflake symbol: ``freeze''.
	}
        \vspace{-0.3cm}
	\label{fig:pipeline}
\end{figure*}
As mentioned above, to enable virtual try-on from unconstrained designs, we propose a novel two-stage pipeline based on Stable Diffusion~\cite{rombach2022high} to disentangle clothing types (\eg, long sleeves) and textures from full-body human image data (Sec.~\ref{sec:two_stage_disentanglement}).
In addition, to further improve the texture quality, we propose to learn photorealistic texture features in a hierarchical and balanced way (Sec.~\ref{sec:multi-level}). 
Finally, to endow the model with the ability to learn non-stationary textures as well as the scale and position of textures on the input garment, we propose to incorporate a positional encoding module in our pipeline (Sec.~\ref{sec:position encoding}).
An overview of our pipeline is shown in Fig.~\ref{fig:pipeline}.

\vspace{2mm}
\noindent \textbf{Definitions of Additional Symbols.} 
In addition to the symbols defined in Sec.~\ref{sec:background}, we have:
\begin{itemize}
    \item $\mathbf{I^m}$: the clothing-agnostic version of $I$, which is generated by i) masking the whole bounding box area of a garment in $I$ and ii) copy-pasting the hair area of $I$ back on top. These can be easily achieved using the parsing map of $I$.
    Note that for lower garments, we ensure the bottom edges of their bounding boxes always go down to the shoes to avoid the leakage of garment length (\eg, short skirts).
    \item $\mathbf{P(I)}$, $\mathbf{P(I^m)}$: the human parsing images of $I$ and $I^m$.
    \item $\mathbf{D(I)}$: the densepose image of $I$ generated by applying readily available methods \cite{wu2019detectron2} on the SHHQ dataset \cite{fu2022styleganhuman}.
    \item $\mathbf{P(I^m, C_s)}$: the output parsing map generated by inpainting $P(I^m)$ with the guidance of $C_{s}^{text}$ or $C_{s}^{img}$.
    \item $\mathbf{z_{t}^{p}}$ , $\mathbf{z_{t}^{i}}$: the latent variable of output parsing $P(I^m, C_s)$ and output human image $\hat{I}$.
\end{itemize}


\subsection{Two-Stage Style and Texture Disentanglement}
\label{sec:two_stage_disentanglement}

To make virtual try-on effective with unconstrained designs, we need to disentangle their style and texture so that the style of the texture-conditioning image $C_{t}^{img}$ (\eg, trousers) and the texture of the style conditions $C_{s}^{img}$/ $C_{s}^{text}$ do not affect the results.
This is a challenging task for Stable Diffusion~\cite{rombach2022high}, the foundation model we use, as the embeddings generated by its CLIP encoder contain both style and texture information of the input.
To address this issue, we propose a novel two-stage pipeline for style and texture disentanglement, where the first stage aims to generate a parsing map whose clothing style is determined by $C_{s}^{img}$ or $C_{s}^{text}$, and the second stage adds clothing texture conditioned on $C_{t}^{img}$ to the garment areas of the parsing map.
Note that we only used $C_{t}^{img}$ during training and the learned disentanglement allows for using $C_{t}^{patch}$ during inference.
 

\vspace{2mm}
\noindent\textbf{Stage 1: Parsing-based Style Editing.}
Given the masked clothing-agnostic human parsing image $P(I^m)$ and its corresponding densepose image $D(I)$ where the arm or leg positions of $I$ are provided, we aim to generate an output parsing map $P(I^m, C_s)$ in which the relevant garment region of $P(I^m)$ is edited according to $C_{s}^{img}$ or $C_{s}^{text}$.
Specifically, we freeze the encoder $\mathcal{E}$, decoder $\mathcal{D}$ and the CLIP encoder of Stable Diffusion and finetune its U-Net component using the inputs mentioned above. Note that we included an MLP layer between the CLIP encoder and U-Net as an adaptor to map the CLIP features to the garment style space.
Following~\cite{rombach2022high}, we pass $P(I^m)$ and $D(I)$ through the encoder $\mathcal{E}$ and concatenate their embeddings with latent variable $z_{t}^{p}$ along the channel dimension as input to the U-Net and have:
\begin{equation}
\begin{aligned}
\epsilon_{t} =  \epsilon_{\theta}([\mathcal E (D(I)),\mathcal E(P(I^{m})), z_{t}],t,C_{s})
\end{aligned}
\label{eq:stage1}
\end{equation}
where $C_{s} = \mathrm{MLP}([C_{s}^{text},C_{s}^{img}])$, the sizes of $C_{s}^{text}$ and $C_{s}^{img}$ are $1 \times 768$. 
During training, either $C_{s}^{text}$ or $C_{s}^{img}$ is set to 0 to ensure that only one style signal is used as input.

\vspace{1mm}
\noindent \textit{Remark.}
Our parsing-based style editing module inherently enables sequential editing by simply replacing $P(I)$ with the generated $P(I^m, C_s)$ and running the module again.
Additionally, in the special case when the source human is wearing separate top and bottom garments, and the target garment style is a dress or jumpsuit, the mask should cover all garment parts to implement the editing properly.

\vspace{2mm}
\noindent \textbf{Stage 2: Style-guided Garment Texture Inpainting.} 
Using the edited human parsing map $P(I^m, C_s)$ obtained from Stage 1 as a style guidance, we aim to inpaint the masked region (mostly garment) of $I_m$ according to an input texture reference $C_{t}^{img}$.
Specifically, we follow~\cite{rombach2022high} and implement the inpainting by injecting the CLIP features extracted from $C_{t}^{img}$ into the U-Net through cross-attention. 
However, a naive application of this approach does not work as the final-block CLIP features are rich in semantics but lack low-level texture, position and scale information.
Addressing this issue, we propose the use of hierarchical and balanced CLIP features (\cref{sec:multi-level}) equipped with position encoding (\cref{sec:position encoding}) to learn photorealistic and non-stationary texture features, respectively:
\begin{equation}
C_{t} = \mathcal P_{e}(\mathrm{MLP}(\mathcal H(C_{t}^{img})))
\label{eq:stage2_feature}
\end{equation}
where $\mathcal H$ denotes the extraction of hierarchical CLIP features, and $\mathcal P_{e}$ denotes the position encoding. 
We finetune a Stable Diffusion~\cite{rombach2022high} model for this stage as well and have:
\begin{equation}
\epsilon_{t} =  \epsilon_{\theta}([\mathcal E (P(I^m, C_s)),\mathcal E(I^m), z_{t}],t,C_{t})
\end{equation}


\subsection{Photorealistic Texture Transfer}
\label{sec:multi-level}
\begin{figure}[t]
	\centering
	\includegraphics[width=0.95\linewidth]{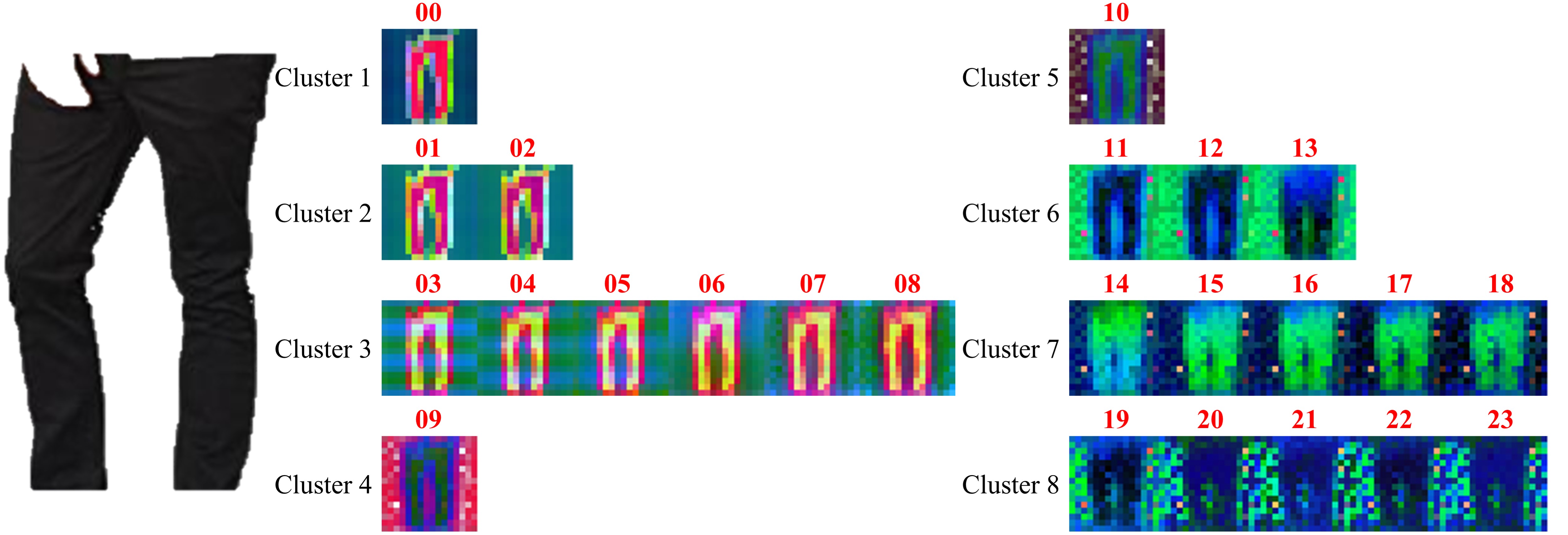}
	\caption{
 Visualization of clustered features from all 24 blocks of CLIP's Vision Transformer. \textcolor{red}{Red}: block id. The higher the block in the hierarchy, the more semantic and less texture information it contains. Please zoom in to see more details.
 }
        \vspace{-0.5cm}
	\label{fig:feature_visual}
\end{figure}

A core challenge in photorealistic texture transfer is determining optimal image features to input into the U-Net. 
A naive solution is to replace the text CLIP features used in previous methods with image CLIP features as images contain richer texture information. 
However, this does not bring significant improvement as we observed that {\it the low-level texture cues are gradually washed away when passing through CLIP blocks} (Fig. \ref{fig:feature_visual}), and the resulting image CLIP features also encapsulate only high-level semantics.
To solve this problem, we propose to counter the wash-away effect and achieve photorealistic texture transfer by using hierarchical and balanced CLIP features.

\vspace{2mm}
\noindent \textbf{Hierarchical Texture Features.} To incorporate low-level texture features, we propose using features extracted from all 24 blocks of CLIP's Vision Transformer. 
To justify this approach, we analyze these features by applying PCA (Principal Component Analysis) to generate 24 feature maps of size $16\times16\times3$.
The features are then classified into eight categories through clustering.
As shown in Fig. \ref{fig:feature_visual}, there are clear differences among the eight feature clusters, indicating that hierarchical texture features provide more thorough information compared to using only the last block. 

\vspace{2mm}

\noindent \textbf{Balancing Texture Features.}
From Fig. \ref{fig:feature_visual}, we observe that the features within each category are similar, while there are significant differences between categories.
In addition, we find that for texture transfer, both low-level features that provide details and high-level features that provide semantic guidance are required.
To minimize redundancy and balance the contributions of different types of feature, we select the shallowest layer feature from each class as the representative feature. 
Using this method, we obtain a representative set of hierarchical CLIP features with dimensions (257$\times$8)$\times$1024. These features are then transformed into (257$\times$8)$\times$768 representations via an MLP layer before being fed into the U-Net.

In this way, we obtain a minimal and balanced set of features $C_t$ that span all levels of CLIP representations, thereby producing photorealistic results.


\subsection{Learning Non-stationary Texture Features}
\label{sec:position encoding}

While CLIP features excel at generating stationary textures like solid colors, they struggle with non-stationary patterns like plaids, inaccurately capturing their scales and color distributions.
To address this, we add a positional encoding layer $\mathcal{P}_e$ before feeding the CLIP features $C_t$ obtained above into the U-Net. 
The added positional information allows for capturing intricate visual details like plaid scales and color variations, markedly improving output quality.
Thus, this enhancement substantially improves the model's ability to represent non-stationary garment textures. 
\section{Experiment}
\begin{figure}[t]
	\centering
	\includegraphics[width=0.98\linewidth]{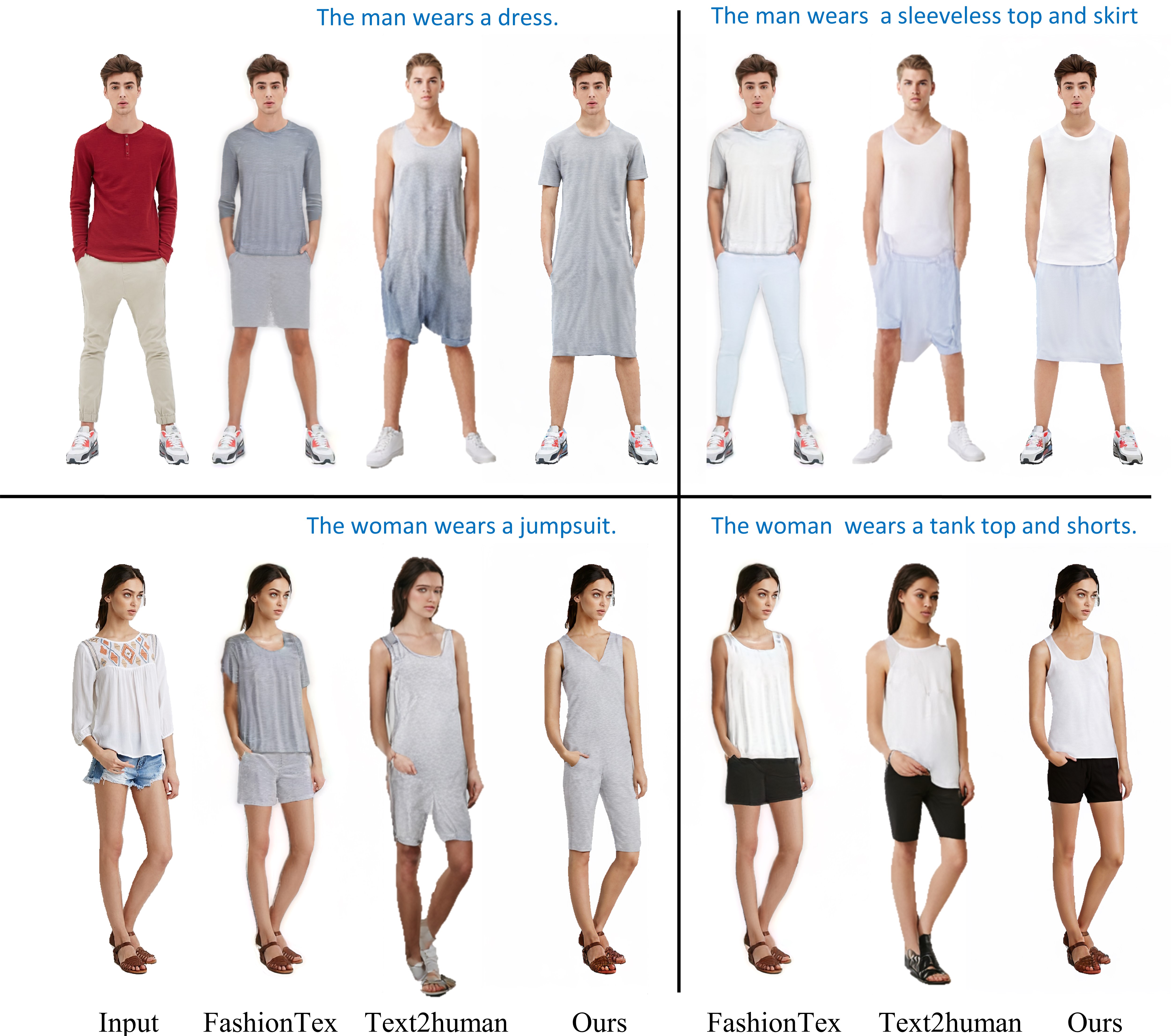}
 \vspace{-0.1cm}
	\caption{Comparison of our model on clothing  style editing. We use a garment image reference for better visualization.
	}
        \vspace{-0.5cm}
	\label{fig:style_comparison}
\end{figure}

\label{sec:experiment}

\subsection{Experimental Setup}

\noindent \textbf{Datasets.} While we evaluate our method on the DeepFashion Multimodal~\cite{liuLQWTcvpr16DeepFashion}, SHHQ~\cite{fu2022styleganhuman} and VITON-HD~\cite{choi2021viton} datasets, we conduct the main experiments on the DeepFashion Multimodal dataset to enable fair comparison, as it serves as the primary benchmark across most state-of-the-art approaches.


\vspace{2mm}
\noindent \textbf{Implementation Details.} 
{\it [Stage 1]} 
For the DeepFashion-Multimodal dataset, we use both text and garment images as style references. For the SHHQ and VITON-HD datasets, which lack text annotations, we exclusively utilize images as inputs. The training and testing resolutions are $512 \times 256$ and $1,024 \times 512$ for DeepFashion-Multimodal and SHHQ, while we use $512 \times 384$ / $1,024 \times 768$ for VITON-HD. 
Each model is fine-tuned from a pre-trained stable diffusion model~\cite{rombach2022high} for 50 epochs with an initial learning rate of $1e^{-5}$.
The classifier-free guidance scale is set to 8 for testing. 
{\it [Stage 2]} We extract clothing from human images and change the background pixel values to 255. 
The training and testing resolutions match Stage 1. 
We fine-tune each model for 100 epochs using the same learning rate as Stage 1. 
The guidance scale is set to 20 for testing. 


\subsection{Comparison with SOTA Methods}

\noindent \textbf{Style Prediction Accuracy.}
To facilitate a fair comparison, we leverage text descriptions $C_{s}^{text}$ as style guidance and evaluate all methods on 6 common clothing styles: ``dress'', ``jumpsuit'', ``short sleeve top and long pants'', ``long sleeve top and shorts'', ``sleeveless top and a skirt''. 
As shown qualitatively in Fig.~\ref{fig:style_comparison} and quantitatively in Table~\ref{tab:style_comparison}, our method achieves significantly higher style prediction accuracy compared to prior arts.
In particular, our approach proficiently generates aligned images from text descriptions, whereas Text2Human~\cite{jiang2022text2human} and FashionTex~\cite{lin2023fashiontex} struggle to produce the desired outcomes from the same textual inputs. 

\begin{figure*}[t]
	\centering
	\includegraphics[width=0.95\linewidth]{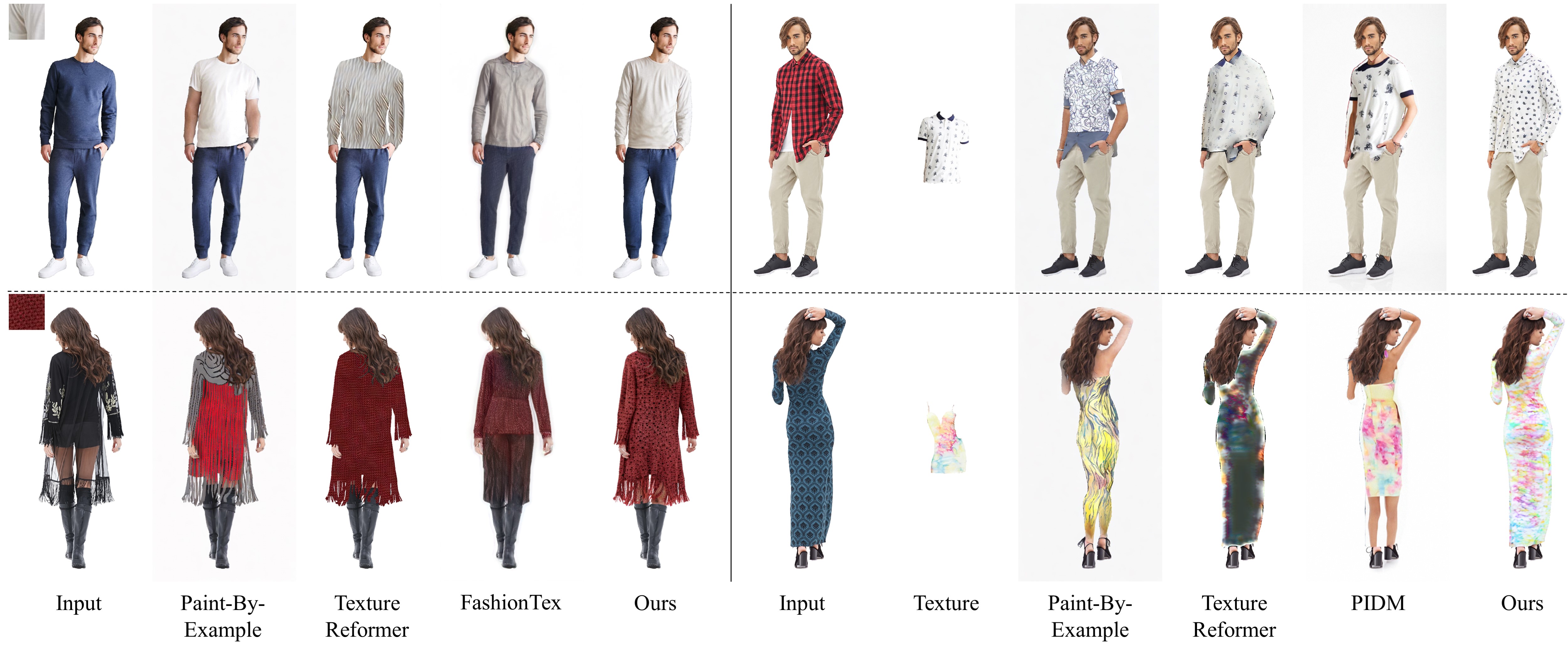}
	\caption{Visual Comparison on garment texture transfer. 
	}
	\label{fig:texture_comparison_visual}
\end{figure*}

\vspace{2mm}
\noindent \textbf{Texture Quality.}
In Table \ref{tab:texture_comparison_quantitative} and Fig.~\ref{fig:texture_comparison_visual}, we provide quantitative and visual comparisons under two different types of texture input $C_t^{img}$: garment texture patches and full garment images. 
Quantitatively, our method outperforms all SOTA methods by a large margin.
Qualitatively, when using garment texture patches as $C_t^{img}$, Texture Reformer~\cite{wang2022texture} and FashionTex \cite{lin2023fashiontex} results exhibit high texture similarity to $C_t^{img}$, yet lack realism as garments.
When using full garment images as $C_t^{img}$, PIDM~\cite{bhunia2023person} generates more garment-like results, but fails to retain the clothing style of the input.
As for Paint-By-Example, it  yields bad results on both patch and image, possibly because it inherently lacks a focus on clothing editing.
However, our approach not only preserves texture fidelity, but also produces realistic, natural-looking human images in both cases. 


\vspace{2mm}
\noindent \textbf{User Study.}
We ran six user studies with 96 participants across various identities and age groups to objectively evaluate our methods compared to others on style/texture fidelity and image quality. Similar to \cite{zhu2023tryondiffusion}, the percentages of each method being selected as the best one are shown in Table \ref{tab:style_comparison} and Table \ref{tab:User study of Texture}, which shows that our method is chosen as the best method by the majority of participants (larger than 70 $\%$) in all cases. More details about the user study are included in the supplementary material.


\begin{table}
  \centering

\begin{tabular}{|c|c|c|c|}
\hline Methods & Accuracy  $\uparrow$ & M $\uparrow$ & R $\uparrow$  \\
\hline 
FashionTex \cite{lin2023fashiontex}& 82.75$\%$ & 3.47$\%$ & 8.33$\%$ \\
Text2Human \cite{jiang2022text2human}& 88.87$\%$ & 7.64$\%$ &9.73$\%$ \\
\hline
\bf Ours & \textbf{92.35}$\boldsymbol{\%}$ & \textbf{88.89}$\boldsymbol{\%}$ & \textbf{81.94}$\boldsymbol{\%}$ \\
\hline

\end{tabular}
\caption{ Comparison for clothing style editing. `Accuracy': The
quantitative comparison of whether the model succeeds in getting
the target cloth type.  `M' and `R':
Two user studies to objectively evaluate our methods compared to
others on style fidelity and image naturalness. }
    \vspace{-0.2cm}
  \label{tab:style_comparison}
\end{table}

\begin{table}
  \centering
  
\setlength\tabcolsep{3pt}
\begin{tabular}{|c|cc|cc|}
\hline  & \multicolumn{2}{|c|}{ Texture patch } & \multicolumn{2}{c|}{Garment } \\
\hline Methods & FID $\downarrow$ & KID $\downarrow$ & FID $\downarrow$ & KID $\downarrow$ \\
\hline Texture Reformer\cite{wang2022texture} & 28.43 & 5.74 & 26.99 & 16.02 \\
Paint-by-Example \cite{yang2023paint}& 28.73 & 6.33 & 27.47 & 5.89 \\
PIDM\cite{bhunia2023person} & -- & -- & 23.90 & 4.40 \\
FashionTex \cite{lin2023fashiontex}& 30.11 & 10.99 & -- & -- \\
\hline
\bf Ours & \textbf{21.25} & \textbf{0.22} & \textbf{22.41}& \textbf{0.81} \\
\hline

\end{tabular}
\caption{ Quantitative Comparisons for garment texture transfer task. The KID is scaled by 1000 following \cite{karras2020training}. }
    \vspace{-0.4cm}
  \label{tab:texture_comparison_quantitative}
\end{table}




\subsection{Generalization across Datasets and Styles}

\begin{figure*}[t]
	\centering
	\includegraphics[width=0.95\linewidth]{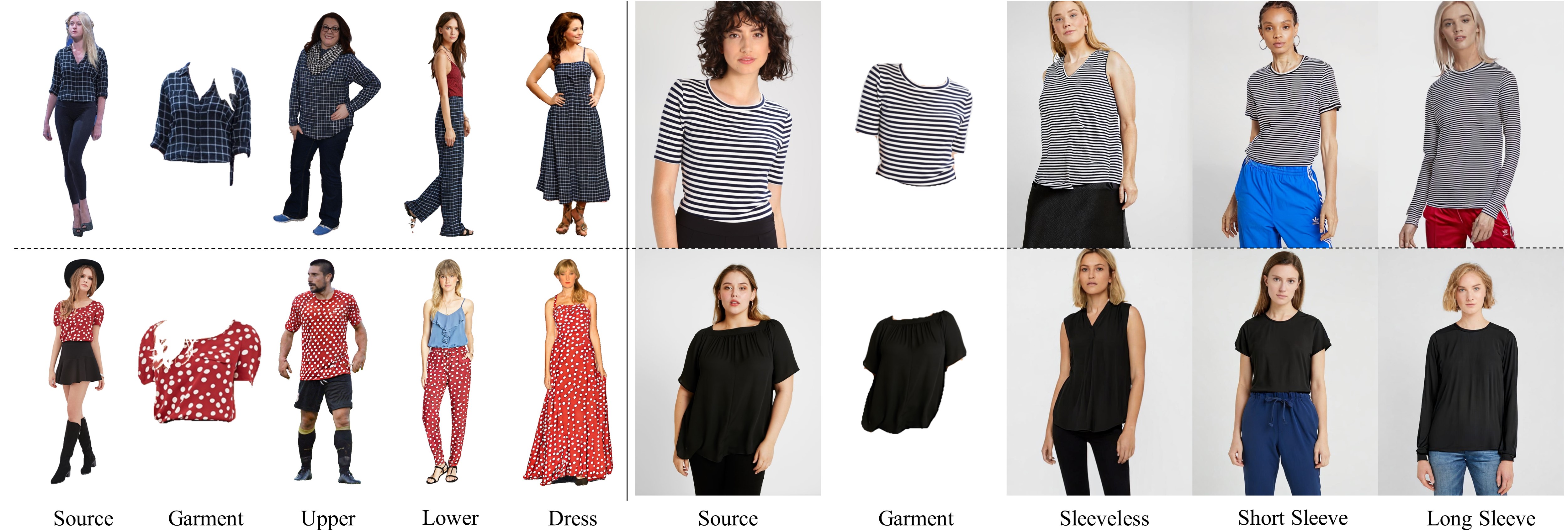}
	\caption{Qualitative results for texture transfer on SHHQ and VITON datasets.
	}
        \vspace{-0.5cm}
	\label{fig:SHHQ_VITON}
\end{figure*}

To study the generalization ability of our method across datasets and styles, Fig.~\ref{fig:SHHQ_VITON} presents our results on the SHHQ and VITON-HD datasets with the same texture guidance $C_{t}^{img}$ but different styles $C_{s}^{text}$. 
For SHHQ, we generate textures on diverse clothing styles, including upperwear, lowerwear, and dresses. 
For VITON-HD, we generate textures on sleeveless, short-sleeved, and long-sleeved garments. 
The highly accurate and visually pleasing results demonstrate that our method is highly generalizable across datasets and clothing styles.

\begin{table}
  \centering
  
\setlength\tabcolsep{2.2pt}
\begin{tabular}{|c|cc|cc|}
\hline  & \multicolumn{2}{|c|}{ Texture patch } & \multicolumn{2}{c|}{Garment } \\
\hline Methods & M $\uparrow$  & R $\uparrow$ & M $\uparrow$ & R $\uparrow$ \\
\hline Texture Reformer\cite{wang2022texture} & 5.73$\%$ & 1.56$\%$ & 5.36$\%$ & 2.98$\%$ \\
Paint-by-Example \cite{yang2023paint}& 5.21$\%$ & 6.76$\%$ & 0$\%$ & 4.17$\%$ \\
PIDM\cite{bhunia2023person} & -- & -- & 23.21$\%$ & 16.66$\%$ \\
FashionTex \cite{lin2023fashiontex}& 1.56$\%$ & 3.13$\%$ & -- & -- \\
\hline
\bf Ours & \textbf{87.5$\boldsymbol{\%}$} & \textbf{88.55$\boldsymbol{\%}$} & \textbf{71.43$\boldsymbol{\%}$}& \textbf{76.19$\boldsymbol{\%}$} \\
\hline

\end{tabular}
\caption{ User studies
to objectively evaluate our methods compared to others at texture fidelity and image naturalness.  }
   \vspace{-0.5cm}
  \label{tab:User study of Texture}
  
\end{table}

\subsection{Ablation Study}

\subsubsection{Effectiveness of Two-stage Disentanglement}

\begin{figure}[t]
	\centering
	\includegraphics[width=0.95\linewidth]{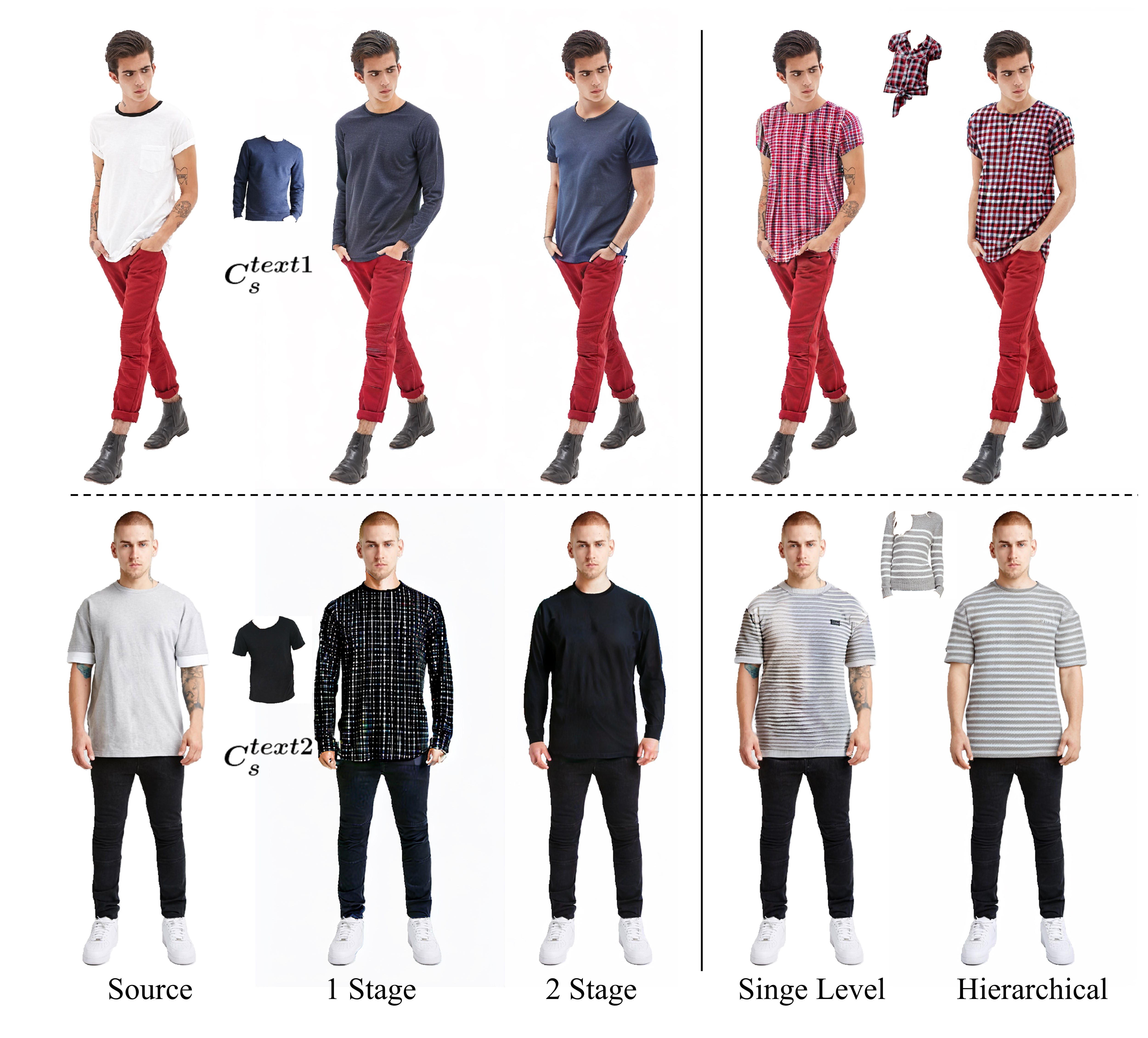}
	\caption{ The ablation studies of our two-stage disentanglement and hierarchical CLIP features. $C_s^{text1}$: {\it ``He wears a short-sleeve T-shirt with floral patterns.''}  $C_s^{text2}$: {\it ``His shirt has long sleeves, cotton fabric and plaid patterns.''} 
	}
	\label{fig:ablation of distanglement and multi level feature}
\end{figure}

To demonstrate the effectiveness of our two-stage style and texture disentanglement approach, we compare it to a one-stage variant.
Specifically, this one-stage variant concatenates the CLIP features for text style input $C_s^{text}$ and image texture input $C_t^{img}$, and integrates them into the U-Net of Stable Diffusion using cross-attention.
Importantly, our two-stage approach enables image style input $C_s^{img}$, which is not possible with the one-stage variant due to the lack of paired training data for different $C_s^{img}$, $C_t^{img}$. This limitation significantly restricts the one-stage method's applicability in real-world scenarios.
For a fair comparison in the limited application scenario of text style input, we remove the hierarchical and balanced feature extraction, and position encoding from our method and use only the last-block CLIP feature of $C_t^{img}$.
As shown in Fig. \ref{fig:ablation of distanglement and multi level feature}, the one-stage model can i) accurately capture textures but be influenced by the style of the reference texture garment when extracting style information; and ii) sometimes capture the correct style but suffer from texture noise mentioned in the text (\eg, plaid patterns). In contrast, our two-stage model can perfectly decouple style and texture and produce highly controllable and realistic results.


\vspace{-0.3cm}
\subsubsection{Effectiveness of Improved CLIP Features}

To demonstrate the effectiveness of our hierarchical and balanced CLIP feature extraction, in Table \ref{tab:ablation-multi_level}, we present a quantitative comparison of our algorithm using CLIP features from different categories. `
Specifically, `Single CLIP feature'' refers to using only the image CLIP feature, while ``layers 1 -- $n$'' indicates concatenating the shallowest feature of the first to $n^{th}$ category, forming a composite feature (Sec.~\ref{sec:multi-level}). Experimental results show that using multi-layer features significantly outperforms single-layer features, with using  shallowest feature of all 8 categories yielding the best results. As Fig. \ref{fig:ablation of distanglement and multi level feature} shows, compared to our results, garments generated using only the Single CLIP feature do not closely match the provided texture. In contrast, our composite CLIP feature enables garment synthesis that better preserves both global style and fine texture details.





\vspace{-3mm}
\subsubsection{Effectiveness of Position Encoding}

\begin{figure}[t]
	\centering
	\includegraphics[width=0.95\linewidth]{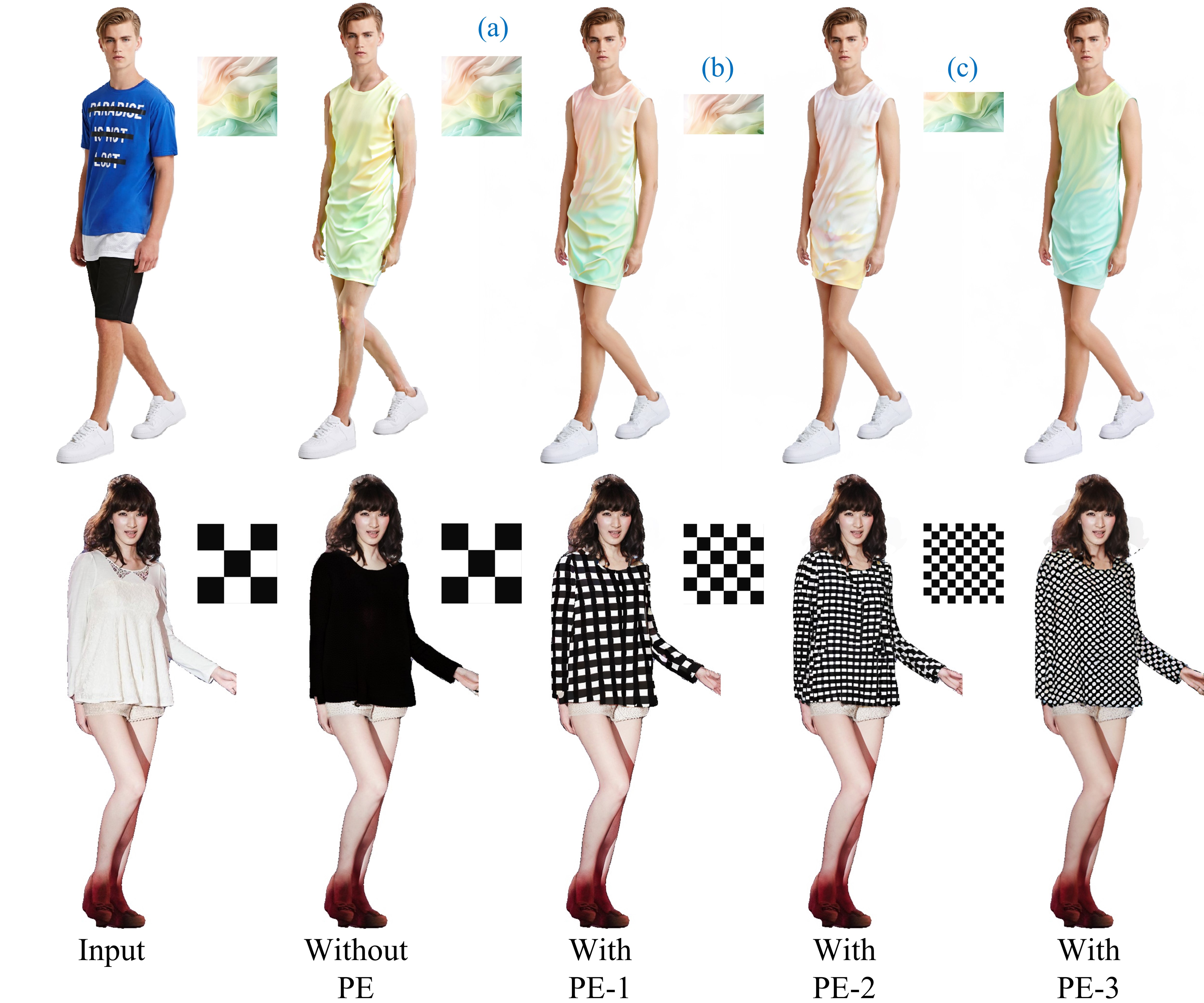}
	\caption{The effectiveness of position encoding, where (b) and (c) are the upper part and lower part of (a) respectively.
	}
        \vspace{-0.5cm}
	\label{fig:ablation_PE}
\end{figure}

In Fig. \ref{fig:ablation_PE}, we conduct three experiments to analyze the effect of position encoding, including: i) without position encoding; ii) input different cropped areas from a single reference garment $C_{t}^{img}$ with non-stationary textures; and iii) input $C_{t}^{img}$ with the same pattern but at different scales. 
The results show position encoding significantly enhances the correlation between the generated garments and provided textures. Specifically, it not only i) preserves the texture distribution (\eg, non-stationary) of $C_{t}^{img}$ effectively, but also ii) enables generating results at different scales, introducing greater flexibility. 
Overall, position encoding is crucial for establishing spatial correspondence between input texture and output garments, enabling virtual try-on and fashion design that adapt to various scenarios.


\begin{table}
  \centering
  \begin{tabular}{|c|c|c|}
    \hline 
    Method & FID $\downarrow$ & KID  $\downarrow$\\
    \hline 
    Single CLIP feature & 24.57& 4.57\\
    layers 1 -- 2 & 22.87& 0.70\\
    layers 1 -- 4& 22.99& 1.04\\
    layers 1 -- 6& 23.06&1.34\\

   \bf Ours (1 -- 8) & \textbf{22.25}&\textbf{0.22}\\
    \hline 
  \end{tabular}
  \caption{Effectiveness of hierarchical and balanced CLIP features. }
  \vspace{-0.5cm}
  \label{tab:ablation-multi_level}
\end{table}

\subsection{Applications}

\begin{figure}[t]
	\centering
	\includegraphics[width=0.95\linewidth]{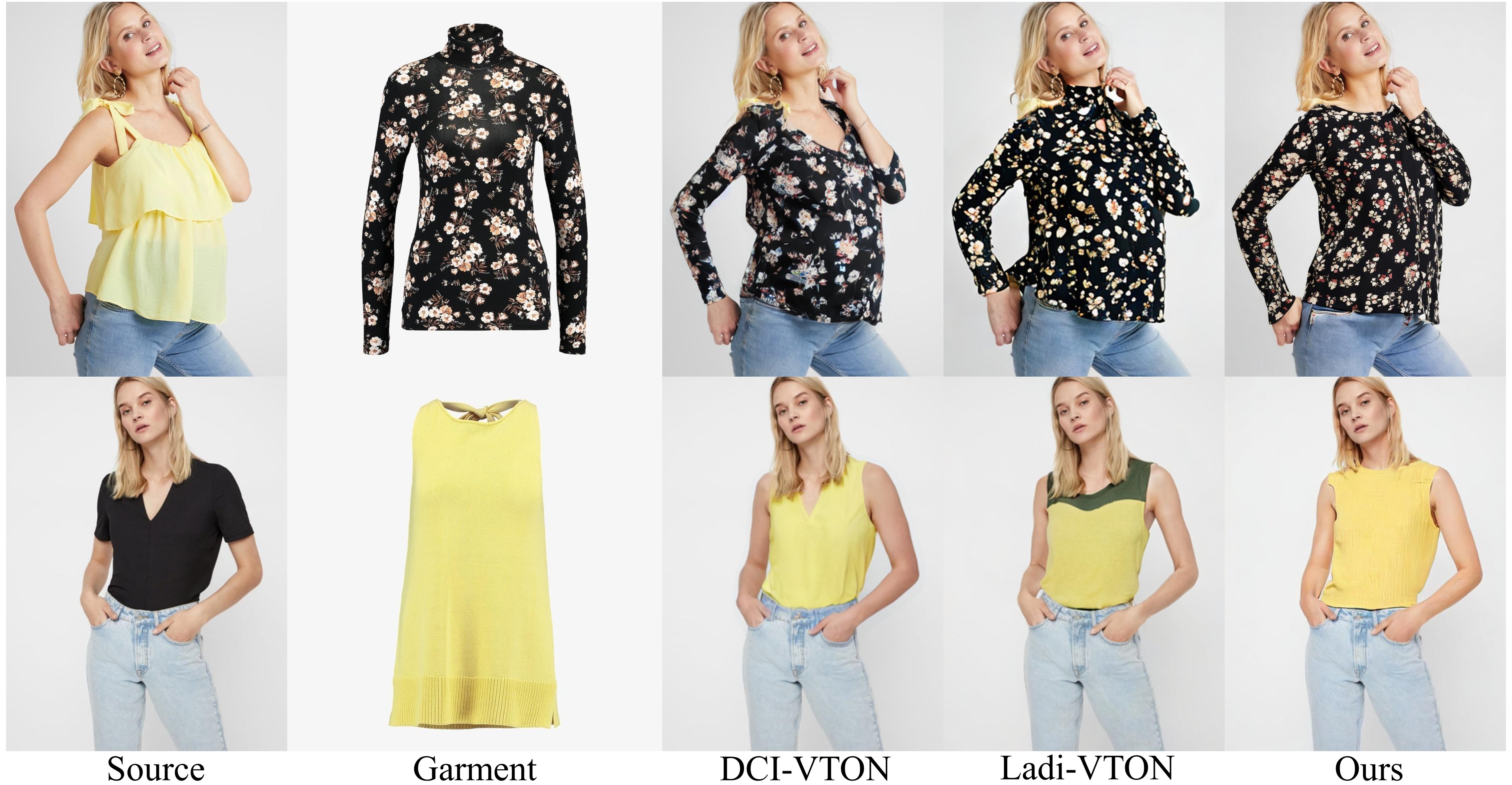}
	\caption{The comparison of virtual try-on based on in-shop cloth.
	}
	\label{fig:try_on}
\end{figure}

\noindent\textbf{In-shop Virtual Try-on.} We qualitatively compare our virtual try-on results to state-of-the-art methods using in-shop clothing images. 
As Fig. \ref{fig:try_on} shows, our approach achieves comparable visual realism and clothing accuracy to recent algorithms, as evidenced by the realistic rendering and natural fit of the virtually dressed person. This demonstrates that our method can synthesize high-fidelity images of in-shop clothing on people using only a single reference image.

\begin{figure}[t]
	\centering
	\includegraphics[width=0.95\linewidth]{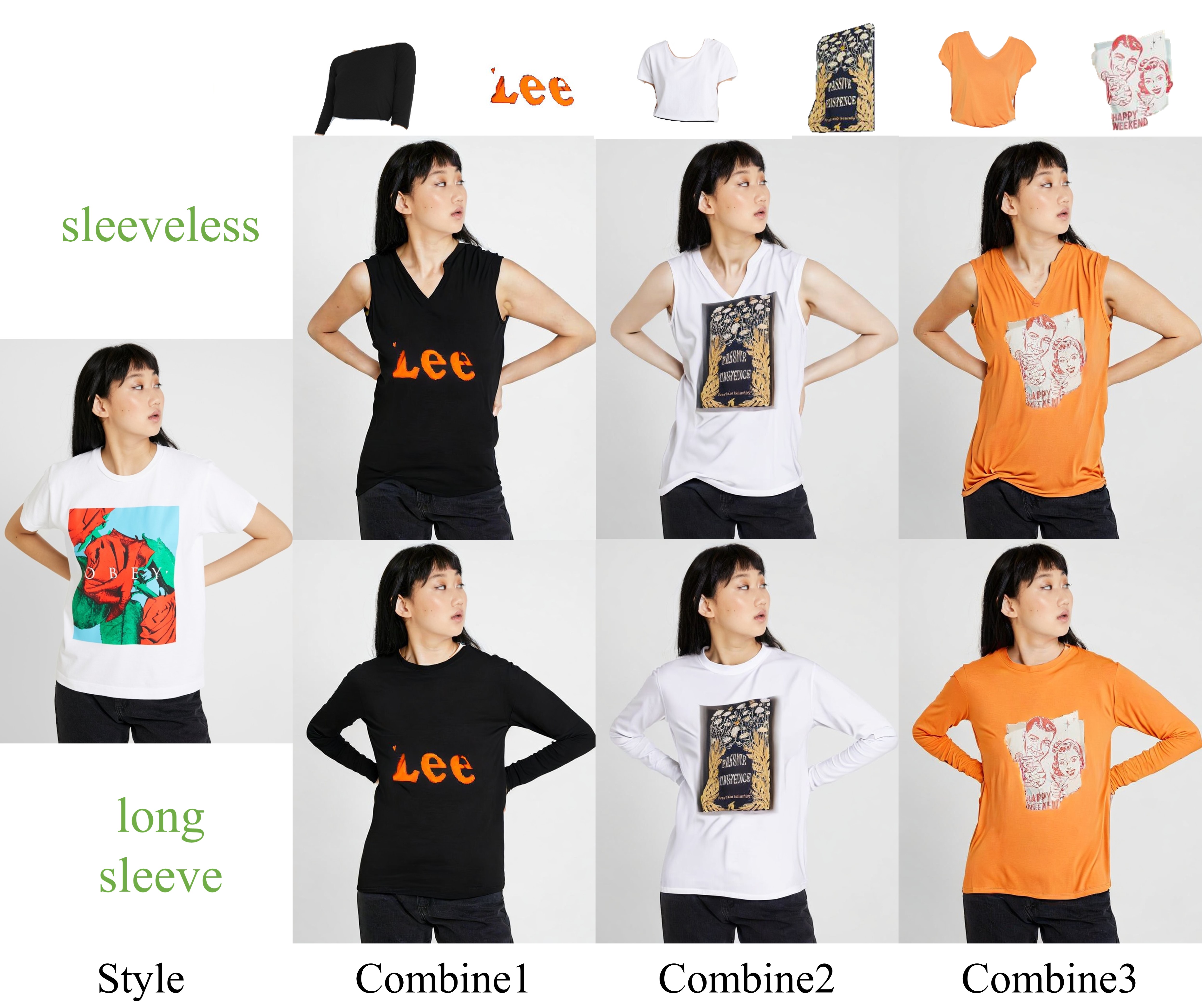}
	\caption{Fashion design results under different conditions including style, texture, and design elements (\eg, logo).
	}
        \vspace{-0.5cm}
	\label{fig:fashion_design}
\end{figure}

\vspace{-0.004cm}
\noindent\textbf{Fashion Design.}
Beyond virtual try-on, our method can also assist fashion design. As shown in Fig. \ref{fig:fashion_design}, given a text prompt specifying the clothing style, a base texture, and a logo, our approach can generate realistic images of a person wearing the described garment. This demonstrates the versatility of our algorithm to not only virtually dress people in existing clothing, but also create new outfit designs from scratch. Our system has the potential to serve as an inspirational tool for fashion designers by instantly visualizing creative concepts.

\vspace{-0.1cm}
\section{Conclusion}
\label{sec:conclusion}

In conclusion, we have introduced the novel task of virtual try-on from unconstrained fashion designs (ucVTON) to enable flexible and photorealistic synthesis of personalized composite clothing. Our key technical contributions include a two-stage disentanglement pipeline to explicitly separate style and texture when using full garment images as complex entangled conditions;a novel hierarchical and balanced CLIP feature extraction module
and position encoding to represent non-stationary textures for high-fidelity synthesis, which significantly expands the diversity of allowable style and texture conditions compared to prior arts. Extensive experiments demonstrate the superiority of our approach in photorealism, personalization, and fine-grained controllability. The flexible mixing and matching of styles and textures enabled by our work brings VTON to a new level that benefits various real-world applications from online shopping to fashion design. 


{
    \small
    \bibliographystyle{ieeenat_fullname}
    \bibliography{main}

\begin{thebibliography}{42}
\providecommand{\natexlab}[1]{#1}
\providecommand{\url}[1]{\texttt{#1}}
\expandafter\ifx\csname urlstyle\endcsname\relax
  \providecommand{\doi}[1]{doi: #1}\else
  \providecommand{\doi}{doi: \begingroup \urlstyle{rm}\Url}\fi

\bibitem[Alaluf et~al.(2022)Alaluf, Tov, Mokady, Gal, and Bermano]{alaluf2022hyperstyle}
Yuval Alaluf, Omer Tov, Ron Mokady, Rinon Gal, and Amit Bermano.
\newblock Hyperstyle: Stylegan inversion with hypernetworks for real image editing.
\newblock In \emph{Proceedings of the IEEE/CVF conference on computer Vision and pattern recognition}, pages 18511--18521, 2022.

\bibitem[Bai et~al.(2022)Bai, Zhou, Li, Zhou, and Yang]{bai2022single}
Shuai Bai, Huiling Zhou, Zhikang Li, Chang Zhou, and Hongxia Yang.
\newblock Single stage virtual try-on via deformable attention flows.
\newblock In \emph{European Conference on Computer Vision}, pages 409--425. Springer, 2022.

\bibitem[Baldrati et~al.(2023)Baldrati, Morelli, Cartella, Cornia, Bertini, and Cucchiara]{baldrati2023multimodal}
Alberto Baldrati, Davide Morelli, Giuseppe Cartella, Marcella Cornia, Marco Bertini, and Rita Cucchiara.
\newblock Multimodal garment designer: Human-centric latent diffusion models for fashion image editing.
\newblock \emph{arXiv preprint arXiv:2304.02051}, 2023.

\bibitem[Bhunia et~al.(2023)Bhunia, Khan, Cholakkal, Anwer, Laaksonen, Shah, and Khan]{bhunia2023person}
Ankan~Kumar Bhunia, Salman Khan, Hisham Cholakkal, Rao~Muhammad Anwer, Jorma Laaksonen, Mubarak Shah, and Fahad~Shahbaz Khan.
\newblock Person image synthesis via denoising diffusion model.
\newblock In \emph{Proceedings of the IEEE/CVF Conference on Computer Vision and Pattern Recognition}, pages 5968--5976, 2023.

\bibitem[Chen et~al.(2020)Chen, Tian, Li, Wu, King, Chen, Hsieh, and Xu]{chen2020tailorgan}
Lele Chen, Justin Tian, Guo Li, Cheng-Haw Wu, Erh-Kan King, Kuan-Ting Chen, Shao-Hang Hsieh, and Chenliang Xu.
\newblock Tailorgan: making user-defined fashion designs.
\newblock In \emph{Proceedings of the IEEE/CVF Winter Conference on Applications of Computer Vision}, pages 3241--3250, 2020.

\bibitem[Choi et~al.(2021)Choi, Park, Lee, and Choo]{choi2021viton}
Seunghwan Choi, Sunghyun Park, Minsoo Lee, and Jaegul Choo.
\newblock Viton-hd: High-resolution virtual try-on via misalignment-aware normalization.
\newblock In \emph{Proceedings of the IEEE/CVF conference on computer vision and pattern recognition}, pages 14131--14140, 2021.

\bibitem[Dong et~al.(2019)Dong, Liang, Shen, Wang, Lai, Zhu, Hu, and Yin]{dong2019towards}
Haoye Dong, Xiaodan Liang, Xiaohui Shen, Bochao Wang, Hanjiang Lai, Jia Zhu, Zhiting Hu, and Jian Yin.
\newblock Towards multi-pose guided virtual try-on network.
\newblock In \emph{Proceedings of the IEEE/CVF international conference on computer vision}, pages 9026--9035, 2019.

\bibitem[Fu et~al.(2022)Fu, Li, Jiang, Lin, Qian, Loy, Wu, and Liu]{fu2022styleganhuman}
Jianglin Fu, Shikai Li, Yuming Jiang, Kwan-Yee Lin, Chen Qian, Chen-Change Loy, Wayne Wu, and Ziwei Liu.
\newblock Stylegan-human: A data-centric odyssey of human generation.
\newblock \emph{arXiv preprint}, arXiv:2204.11823, 2022.

\bibitem[Ge et~al.(2021)Ge, Song, Zhang, Ge, Liu, and Luo]{ge2021parser}
Yuying Ge, Yibing Song, Ruimao Zhang, Chongjian Ge, Wei Liu, and Ping Luo.
\newblock Parser-free virtual try-on via distilling appearance flows.
\newblock In \emph{Proceedings of the IEEE/CVF conference on computer vision and pattern recognition}, pages 8485--8493, 2021.

\bibitem[Goel et~al.(2023)Goel, Peruzzo, Jiang, Xu, Sebe, Darrell, Wang, and Shi]{goel2023pair}
Vidit Goel, Elia Peruzzo, Yifan Jiang, Dejia Xu, Nicu Sebe, Trevor Darrell, Zhangyang Wang, and Humphrey Shi.
\newblock Pair-diffusion: Object-level image editing with structure-and-appearance paired diffusion models.
\newblock \emph{arXiv preprint arXiv:2303.17546}, 2023.

\bibitem[Gou et~al.(2023)Gou, Sun, Zhang, Si, Qian, and Zhang]{gou2023taming}
Junhong Gou, Siyu Sun, Jianfu Zhang, Jianlou Si, Chen Qian, and Liqing Zhang.
\newblock Taming the power of diffusion models for high-quality virtual try-on with appearance flow.
\newblock \emph{arXiv preprint arXiv:2308.06101}, 2023.

\bibitem[He et~al.(2022)He, Song, and Xiang]{he2022style}
Sen He, Yi-Zhe Song, and Tao Xiang.
\newblock Style-based global appearance flow for virtual try-on.
\newblock In \emph{Proceedings of the IEEE/CVF Conference on Computer Vision and Pattern Recognition}, pages 3470--3479, 2022.

\bibitem[Ho et~al.(2020)Ho, Jain, and Abbeel]{ho2020denoising}
Jonathan Ho, Ajay Jain, and Pieter Abbeel.
\newblock Denoising diffusion probabilistic models.
\newblock \emph{Advances in neural information processing systems}, 33:\penalty0 6840--6851, 2020.

\bibitem[Huang et~al.(2023)Huang, Chen, Liu, Shen, Zhao, and Zhou]{huang2023composer}
Lianghua Huang, Di Chen, Yu Liu, Yujun Shen, Deli Zhao, and Jingren Zhou.
\newblock Composer: Creative and controllable image synthesis with composable conditions.
\newblock \emph{arXiv preprint arXiv:2302.09778}, 2023.

\bibitem[Jiang et~al.(2022)Jiang, Yang, Qiu, Wu, Loy, and Liu]{jiang2022text2human}
Yuming Jiang, Shuai Yang, Haonan Qiu, Wayne Wu, Chen~Change Loy, and Ziwei Liu.
\newblock Text2human: Text-driven controllable human image generation.
\newblock \emph{ACM Transactions on Graphics (TOG)}, 41\penalty0 (4):\penalty0 1--11, 2022.

\bibitem[Karras et~al.(2020)Karras, Aittala, Hellsten, Laine, Lehtinen, and Aila]{karras2020training}
Tero Karras, Miika Aittala, Janne Hellsten, Samuli Laine, Jaakko Lehtinen, and Timo Aila.
\newblock Training generative adversarial networks with limited data.
\newblock \emph{Advances in neural information processing systems}, 33:\penalty0 12104--12114, 2020.

\bibitem[Kim et~al.(2023)Kim, Lee, Kim, Ha, and Zhu]{kim2023dense}
Yunji Kim, Jiyoung Lee, Jin-Hwa Kim, Jung-Woo Ha, and Jun-Yan Zhu.
\newblock Dense text-to-image generation with attention modulation.
\newblock In \emph{Proceedings of the IEEE/CVF International Conference on Computer Vision}, pages 7701--7711, 2023.

\bibitem[Lee et~al.(2022)Lee, Gu, Park, Choi, and Choo]{lee2022high}
Sangyun Lee, Gyojung Gu, Sunghyun Park, Seunghwan Choi, and Jaegul Choo.
\newblock High-resolution virtual try-on with misalignment and occlusion-handled conditions.
\newblock In \emph{European Conference on Computer Vision}, pages 204--219. Springer, 2022.

\bibitem[Lin et~al.(2023)Lin, Zhao, Ning, Qiu, Wang, and Han]{lin2023fashiontex}
Anran Lin, Nanxuan Zhao, Shuliang Ning, Yuda Qiu, Baoyuan Wang, and Xiaoguang Han.
\newblock Fashiontex: Controllable virtual try-on with text and texture.
\newblock \emph{arXiv preprint arXiv:2305.04451}, 2023.

\bibitem[Liu et~al.(2016)Liu, Luo, Qiu, Wang, and Tang]{liuLQWTcvpr16DeepFashion}
Ziwei Liu, Ping Luo, Shi Qiu, Xiaogang Wang, and Xiaoou Tang.
\newblock Deepfashion: Powering robust clothes recognition and retrieval with rich annotations.
\newblock In \emph{Proceedings of IEEE Conference on Computer Vision and Pattern Recognition (CVPR)}, 2016.

\bibitem[Morelli et~al.(2022)Morelli, Fincato, Cornia, Landi, Cesari, and Cucchiara]{morelli2022dress}
Davide Morelli, Matteo Fincato, Marcella Cornia, Federico Landi, Fabio Cesari, and Rita Cucchiara.
\newblock Dress code: High-resolution multi-category virtual try-on.
\newblock In \emph{Proceedings of the IEEE/CVF Conference on Computer Vision and Pattern Recognition}, pages 2231--2235, 2022.

\bibitem[Morelli et~al.(2023)Morelli, Baldrati, Cartella, Cornia, Bertini, and Cucchiara]{morelli2023ladi}
Davide Morelli, Alberto Baldrati, Giuseppe Cartella, Marcella Cornia, Marco Bertini, and Rita Cucchiara.
\newblock {LaDI-VTON: Latent Diffusion Textual-Inversion Enhanced Virtual Try-On}.
\newblock In \emph{Proceedings of the ACM International Conference on Multimedia}, 2023.

\bibitem[Mou et~al.(2023)Mou, Wang, Xie, Zhang, Qi, Shan, and Qie]{mou2023t2i}
Chong Mou, Xintao Wang, Liangbin Xie, Jian Zhang, Zhongang Qi, Ying Shan, and Xiaohu Qie.
\newblock T2i-adapter: Learning adapters to dig out more controllable ability for text-to-image diffusion models.
\newblock \emph{arXiv preprint arXiv:2302.08453}, 2023.

\bibitem[Patashnik et~al.(2021)Patashnik, Wu, Shechtman, Cohen-Or, and Lischinski]{patashnik2021styleclip}
Or Patashnik, Zongze Wu, Eli Shechtman, Daniel Cohen-Or, and Dani Lischinski.
\newblock Styleclip: Text-driven manipulation of stylegan imagery.
\newblock In \emph{Proceedings of the IEEE/CVF International Conference on Computer Vision}, pages 2085--2094, 2021.

\bibitem[Ren et~al.(2023)Ren, Delbracio, Talebi, Gerig, and Milanfar]{ren2023multiscale}
Mengwei Ren, Mauricio Delbracio, Hossein Talebi, Guido Gerig, and Peyman Milanfar.
\newblock Multiscale structure guided diffusion for image deblurring.
\newblock In \emph{Proceedings of the IEEE/CVF International Conference on Computer Vision}, pages 10721--10733, 2023.

\bibitem[Rombach et~al.(2022)Rombach, Blattmann, Lorenz, Esser, and Ommer]{rombach2022high}
Robin Rombach, Andreas Blattmann, Dominik Lorenz, Patrick Esser, and Bj{\"o}rn Ommer.
\newblock High-resolution image synthesis with latent diffusion models.
\newblock In \emph{Proceedings of the IEEE/CVF conference on computer vision and pattern recognition}, pages 10684--10695, 2022.

\bibitem[Saini et~al.(2023)Saini, Wang, Swaminathan, Jayasundara, He, Gupta, and Shrivastava]{saini2023chop}
Nirat Saini, Hanyu Wang, Archana Swaminathan, Vinoj Jayasundara, Bo He, Kamal Gupta, and Abhinav Shrivastava.
\newblock Chop \& learn: Recognizing and generating object-state compositions.
\newblock In \emph{Proceedings of the IEEE/CVF International Conference on Computer Vision}, pages 20247--20258, 2023.

\bibitem[Song et~al.(2020)Song, Meng, and Ermon]{song2020denoising}
Jiaming Song, Chenlin Meng, and Stefano Ermon.
\newblock Denoising diffusion implicit models.
\newblock \emph{arXiv preprint arXiv:2010.02502}, 2020.

\bibitem[Sun et~al.(2023)Sun, Zhou, He, and Mok]{sun2023sgdiff}
Zhengwentai Sun, Yanghong Zhou, Honghong He, and PY Mok.
\newblock Sgdiff: A style guided diffusion model for fashion synthesis.
\newblock \emph{arXiv preprint arXiv:2308.07605}, 2023.

\bibitem[Vaswani et~al.(2017)Vaswani, Shazeer, Parmar, Uszkoreit, Jones, Gomez, Kaiser, and Polosukhin]{vaswani2017attention}
Ashish Vaswani, Noam Shazeer, Niki Parmar, Jakob Uszkoreit, Llion Jones, Aidan~N Gomez, {\L}ukasz Kaiser, and Illia Polosukhin.
\newblock Attention is all you need.
\newblock \emph{Advances in neural information processing systems}, 30, 2017.

\bibitem[Wang et~al.(2022)Wang, Zhao, Chen, Li, Zuo, Xing, and Lu]{wang2022texture}
Zhizhong Wang, Lei Zhao, Haibo Chen, Ailin Li, Zhiwen Zuo, Wei Xing, and Dongming Lu.
\newblock Texture reformer: towards fast and universal interactive texture transfer.
\newblock In \emph{Proceedings of the AAAI Conference on Artificial Intelligence}, pages 2624--2632, 2022.

\bibitem[Wu et~al.(2023)Wu, Liu, Zhao, Bui, Lin, Zhang, and Chang]{wu2023harnessing}
Qiucheng Wu, Yujian Liu, Handong Zhao, Trung Bui, Zhe Lin, Yang Zhang, and Shiyu Chang.
\newblock Harnessing the spatial-temporal attention of diffusion models for high-fidelity text-to-image synthesis.
\newblock In \emph{Proceedings of the IEEE/CVF International Conference on Computer Vision}, pages 7766--7776, 2023.

\bibitem[Wu et~al.(2019)Wu, Kirillov, Massa, Lo, and Girshick]{wu2019detectron2}
Yuxin Wu, Alexander Kirillov, Francisco Massa, Wan-Yen Lo, and Ross Girshick.
\newblock Detectron2.
\newblock \url{https://github.com/facebookresearch/detectron2}, 2019.

\bibitem[Xia et~al.(2021)Xia, Yang, Xue, and Wu]{xia2021tedigan}
Weihao Xia, Yujiu Yang, Jing-Hao Xue, and Baoyuan Wu.
\newblock Tedigan: Text-guided diverse face image generation and manipulation.
\newblock In \emph{Proceedings of the IEEE/CVF conference on computer vision and pattern recognition}, pages 2256--2265, 2021.

\bibitem[Xiu et~al.(2023)Xiu, Yang, Cao, Tzionas, and Black]{xiu2023econ}
Yuliang Xiu, Jinlong Yang, Xu Cao, Dimitrios Tzionas, and Michael~J Black.
\newblock Econ: Explicit clothed humans optimized via normal integration.
\newblock In \emph{Proceedings of the IEEE/CVF Conference on Computer Vision and Pattern Recognition}, pages 512--523, 2023.

\bibitem[Xue et~al.(2023)Xue, Huang, Sun, Song, and Zhang]{xue2023freestyle}
Han Xue, Zhiwu Huang, Qianru Sun, Li Song, and Wenjun Zhang.
\newblock Freestyle layout-to-image synthesis.
\newblock In \emph{Proceedings of the IEEE/CVF Conference on Computer Vision and Pattern Recognition}, pages 14256--14266, 2023.

\bibitem[Yang et~al.(2023)Yang, Gu, Zhang, Zhang, Chen, Sun, Chen, and Wen]{yang2023paint}
Binxin Yang, Shuyang Gu, Bo Zhang, Ting Zhang, Xuejin Chen, Xiaoyan Sun, Dong Chen, and Fang Wen.
\newblock Paint by example: Exemplar-based image editing with diffusion models.
\newblock In \emph{Proceedings of the IEEE/CVF Conference on Computer Vision and Pattern Recognition}, pages 18381--18391, 2023.

\bibitem[Yang et~al.(2020)Yang, Zhang, Guo, Liu, Zuo, and Luo]{yang2020towards}
Han Yang, Ruimao Zhang, Xiaobao Guo, Wei Liu, Wangmeng Zuo, and Ping Luo.
\newblock Towards photo-realistic virtual try-on by adaptively generating-preserving image content.
\newblock In \emph{Proceedings of the IEEE/CVF conference on computer vision and pattern recognition}, pages 7850--7859, 2020.

\bibitem[Yu et~al.(2019)Yu, Wang, and Xie]{yu2019vtnfp}
Ruiyun Yu, Xiaoqi Wang, and Xiaohui Xie.
\newblock Vtnfp: An image-based virtual try-on network with body and clothing feature preservation.
\newblock In \emph{Proceedings of the IEEE/CVF international conference on computer vision}, pages 10511--10520, 2019.

\bibitem[Zhang et~al.(2023)Zhang, Rao, and Agrawala]{zhang2023adding}
Lvmin Zhang, Anyi Rao, and Maneesh Agrawala.
\newblock Adding conditional control to text-to-image diffusion models.
\newblock In \emph{Proceedings of the IEEE/CVF International Conference on Computer Vision}, pages 3836--3847, 2023.

\bibitem[Zhao et~al.(2023)Zhao, Chen, Chen, Bao, Hao, Yuan, and Wong]{zhao2023uni}
Shihao Zhao, Dongdong Chen, Yen-Chun Chen, Jianmin Bao, Shaozhe Hao, Lu Yuan, and Kwan-Yee~K Wong.
\newblock Uni-controlnet: All-in-one control to text-to-image diffusion models.
\newblock \emph{arXiv preprint arXiv:2305.16322}, 2023.

\bibitem[Zhu et~al.(2023)Zhu, Yang, Zhu, Reda, Chan, Saharia, Norouzi, and Kemelmacher-Shlizerman]{zhu2023tryondiffusion}
Luyang Zhu, Dawei Yang, Tyler Zhu, Fitsum Reda, William Chan, Chitwan Saharia, Mohammad Norouzi, and Ira Kemelmacher-Shlizerman.
\newblock Tryondiffusion: A tale of two unets.
\newblock In \emph{Proceedings of the IEEE/CVF Conference on Computer Vision and Pattern Recognition}, pages 4606--4615, 2023.

\end{thebibliography}
}

\clearpage
\setcounter{page}{1}
\maketitlesupplementary

\begin{figure*}[t]
	\centering
	\includegraphics[width=0.75\linewidth]{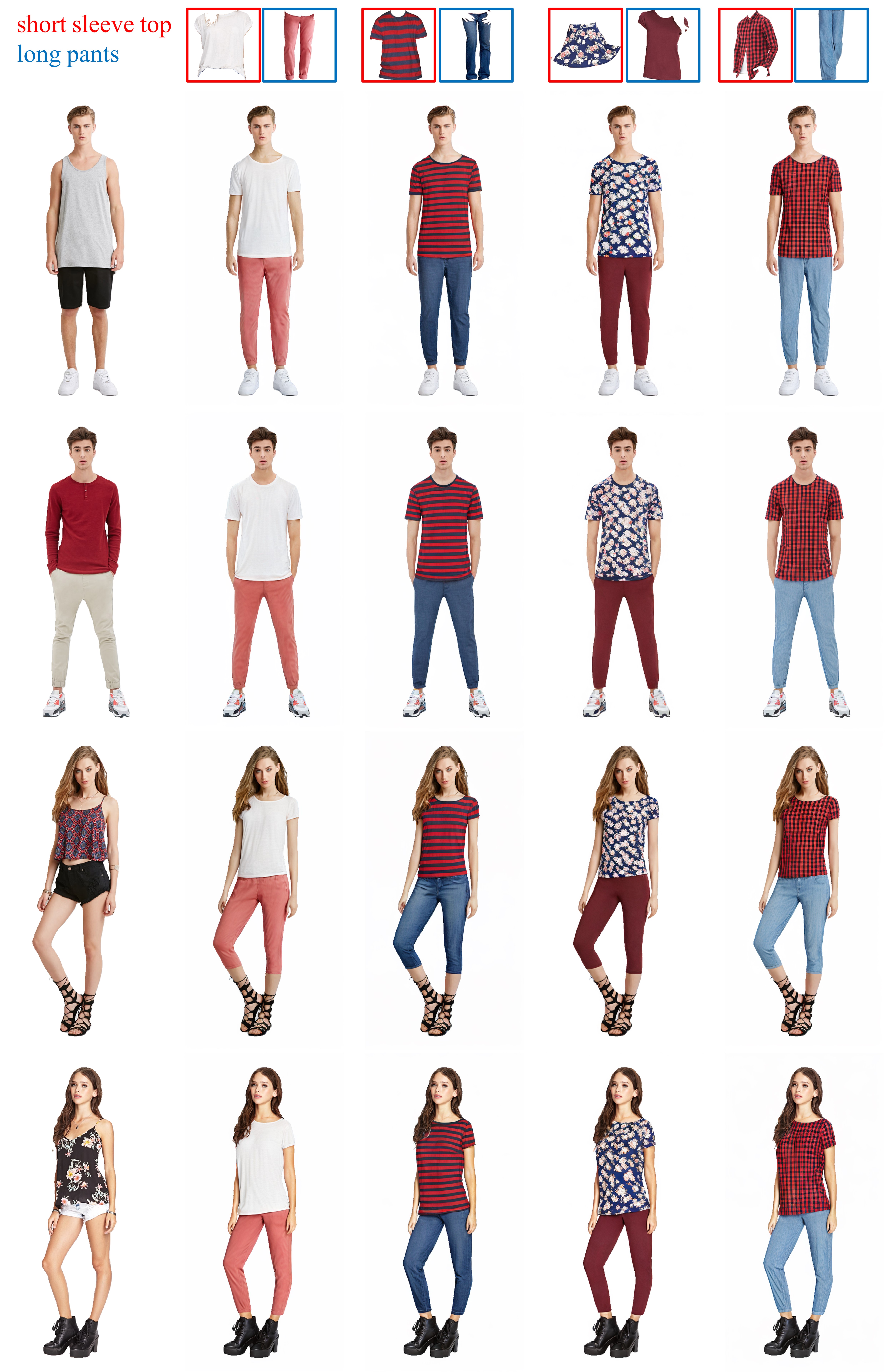}
	\caption{More results of our ucVTON. Red text: style for upper part; Blue text: style for lower paer; Red box: texture for upper part; Blue box: texture for lower part. 
	}
	\label{fig:more1}
\end{figure*}

\begin{figure*}[t]
	\centering
	\includegraphics[width=0.75\linewidth]{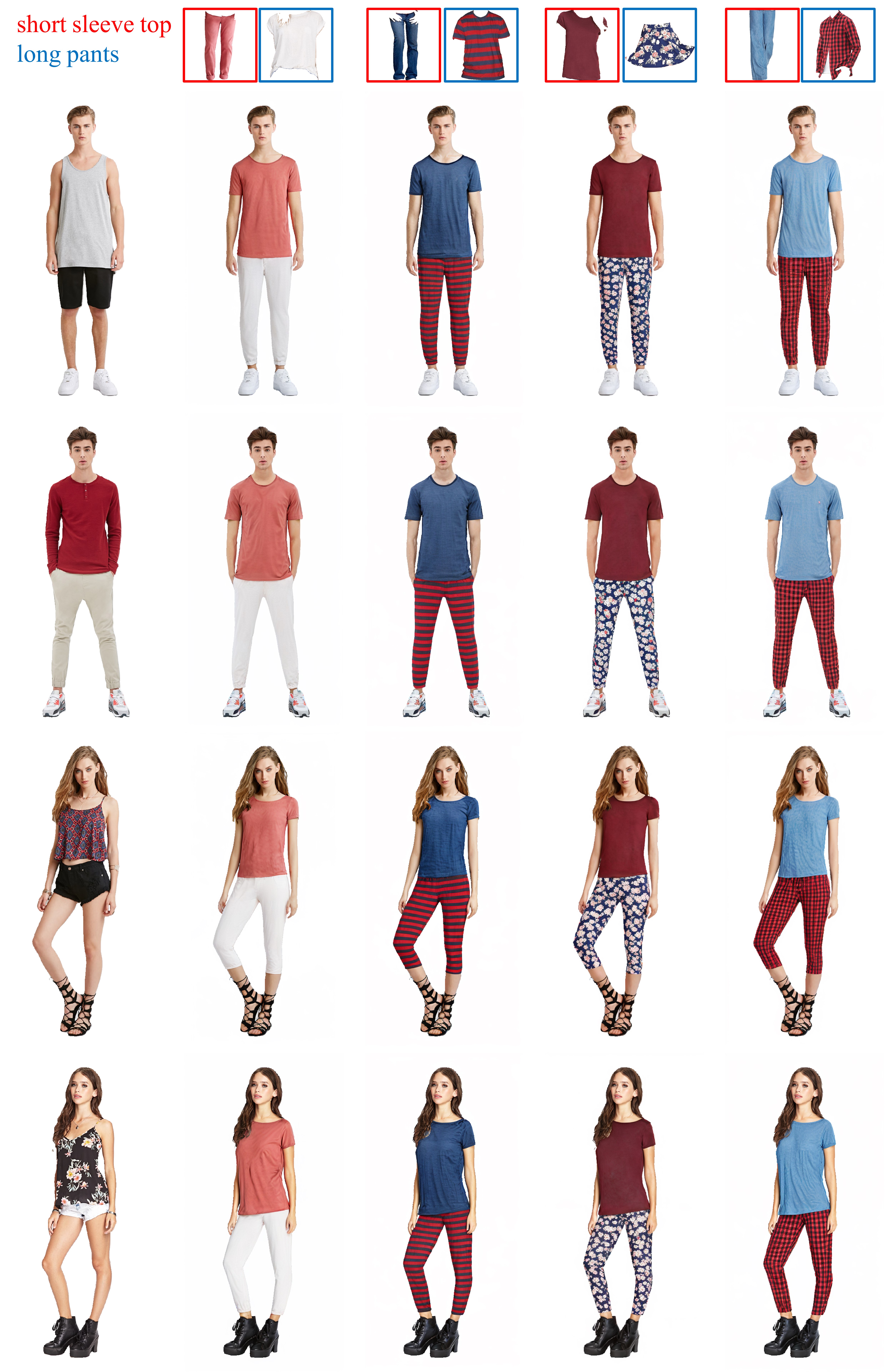}
	\caption{More results of our ucVTON. Red text: style for upper part; Blue text: style for lower paer; Red box: texture for upper part; Blue box: texture for lower part. 
	}
	\label{fig:more2}
\end{figure*}

\begin{figure*}[t]
	\centering
	\includegraphics[width=0.75\linewidth]{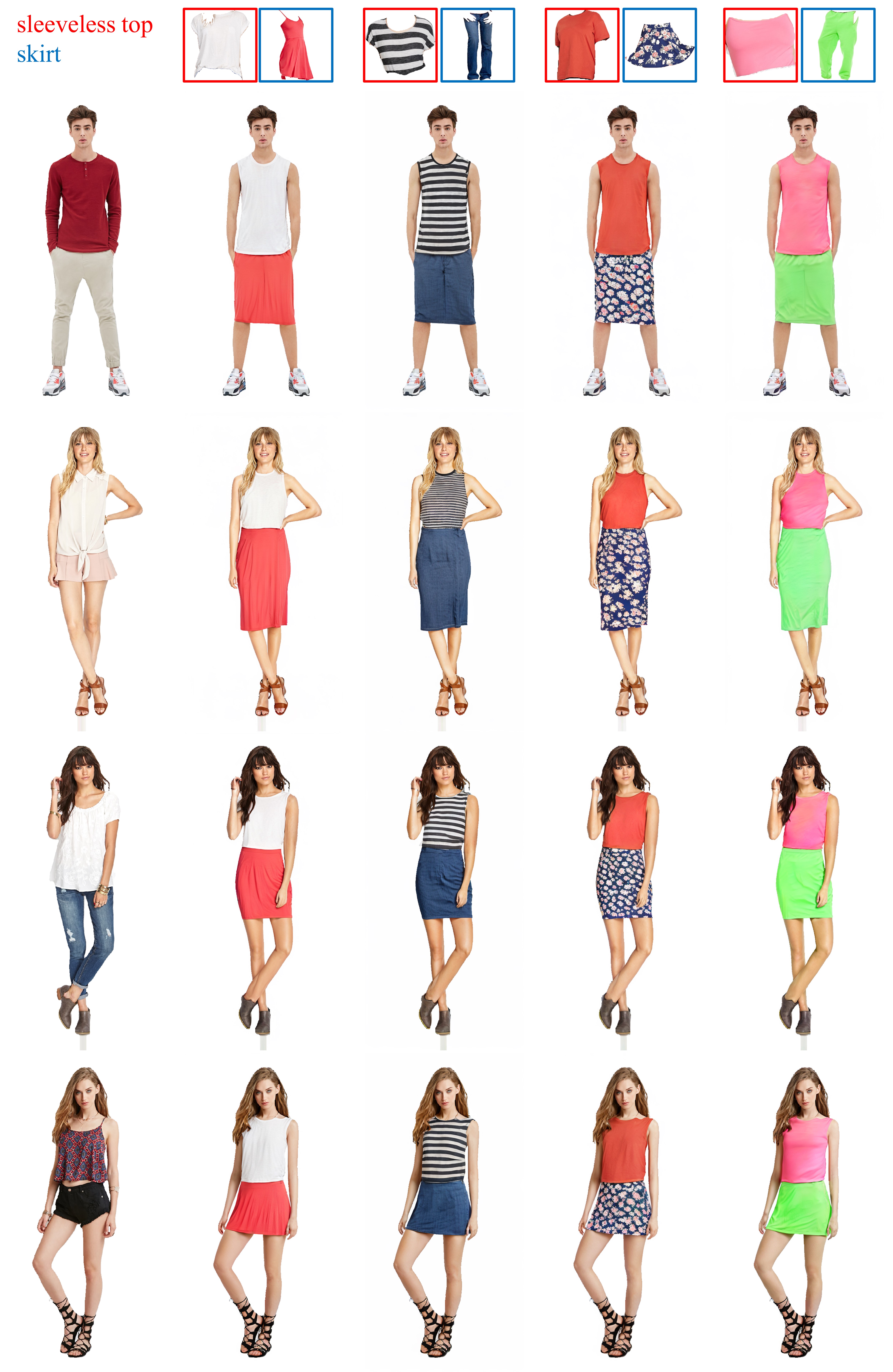}
	\caption{More results of our ucVTON. Red text: style for upper part; Blue text: style for lower paer; Red box: texture for upper part; Blue box: texture for lower part. 
	}
	\label{fig:more3}
\end{figure*}

\begin{figure*}[t]
	\centering
	\includegraphics[width=0.75\linewidth]{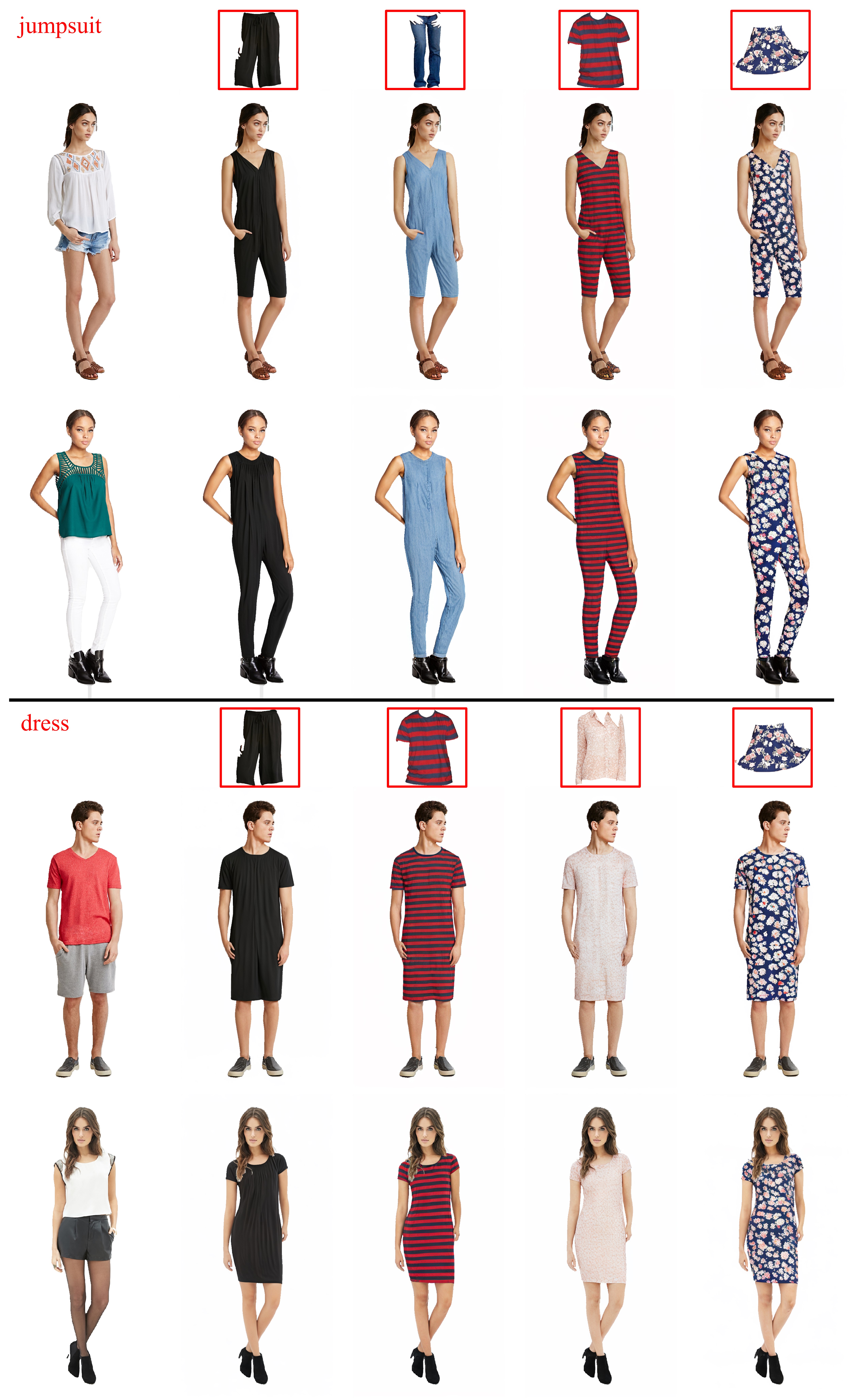}
	\caption{More results of our ucVTON. Red text: style reference;  Red box: texture reference.
	}
	\label{fig:more4}
\end{figure*}

\begin{figure*}[t]
	\centering
	\includegraphics[width=0.95\linewidth]{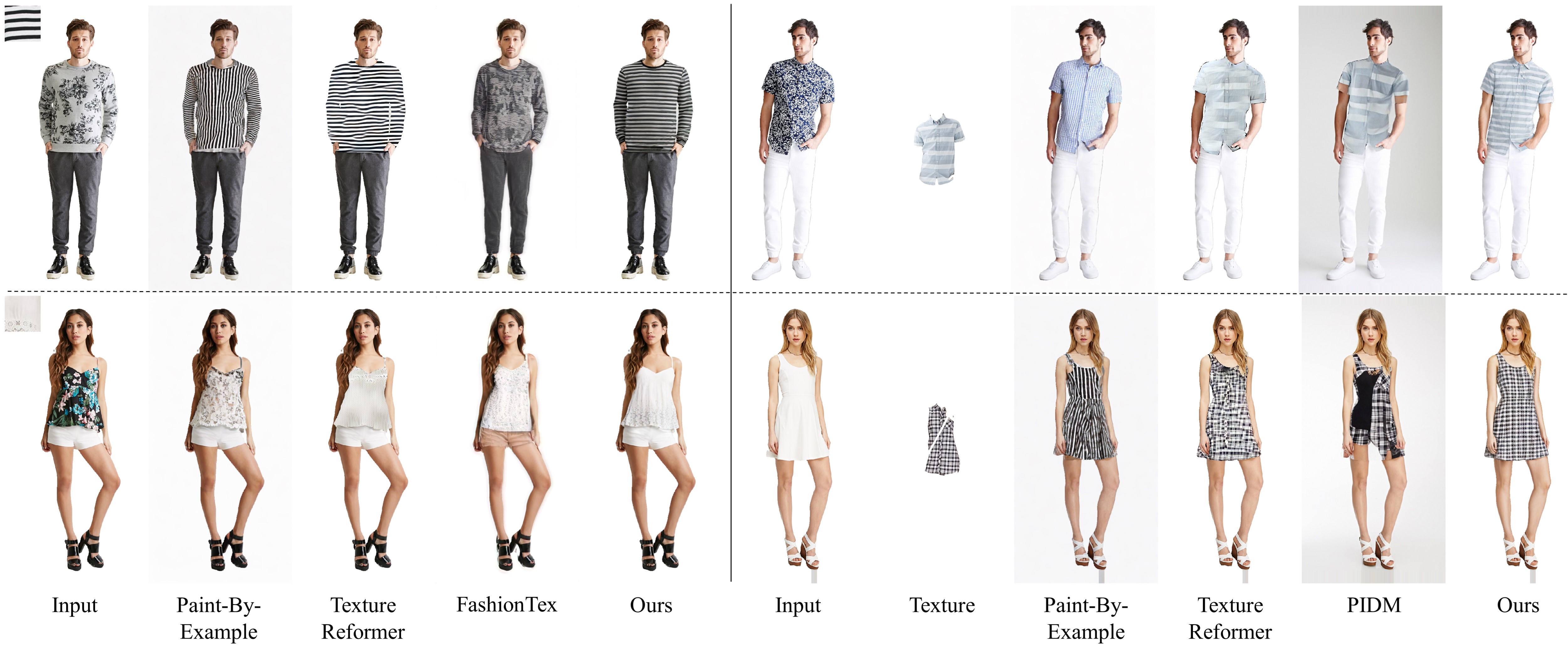}
	\caption{Visual comparison on garment texture transfer. 
	}
	\label{fig:texture_comparison_visual_supp}
\end{figure*}
\begin{figure*}[t]
	\centering
	\includegraphics[width=0.95\linewidth]{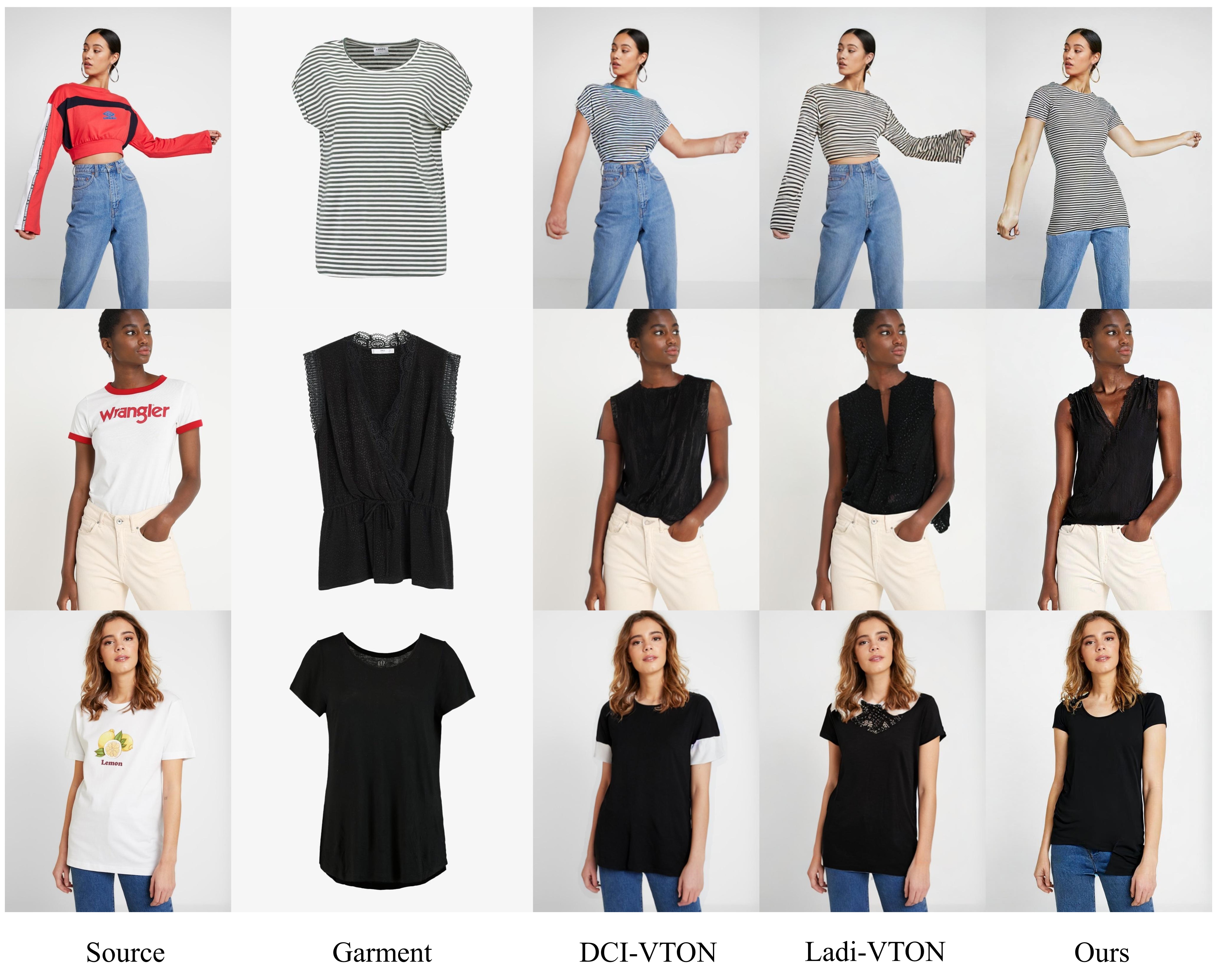}
	\caption{The comparison of virtual try-on based on in-shop cloth.
	}
	\label{fig:try_on_supp}
\end{figure*}

\begin{table*}
  \centering

\begin{tabular}{|c|cc|cc|cc|}
\hline  
\multirow{2}{*}{Methods} & \multicolumn{2}{|c|}{ Style }& \multicolumn{2}{|c|}{ Texture patch } & \multicolumn{2}{c|}{Garment } \\
\cline{2-7}
 & M $\uparrow$ & R $\uparrow$ & M $\uparrow$ & R $\uparrow$  & M $\uparrow$ & R $\uparrow$\\
\hline Texture Reformer\cite{wang2022texture} & -- & -- & 2.58 & 1.64& 1.77 & 0.93 \\
Paint-by-Example \cite{yang2023paint}&-- & -- & 1.88 & 2.41& 1.10 & 1.76 \\
PIDM\cite{bhunia2023person} & -- & -- &-- & -- & 2.56 & 1.98\\
FashionTex \cite{lin2023fashiontex}& 1.24 & 1.66 & 1.72 & 2.14 & -- & --\\
Text2Human \cite{jiang2022text2human}& 1.89 & 1.59 & -- & -- & -- & --\\
\hline
\bf Ours & \textbf{2.86} & \textbf{2.77} & \textbf{3.81}& \textbf{3.82} & \textbf{3.52} & \textbf{3.53}\\
\hline

\end{tabular}
\caption{Additional user studies to objectively compare our methods with others at style fidelity, texture fidelity and image naturalness. }

  \label{tab:user_study}
\end{table*}

  



\section{Preliminaries: Stable Diffusion}
As a SOTA diffusion model, Stable Diffusion~\cite{rombach2022high} consists of an autoencoder $\mathcal{A}$ containing encoder $\mathcal{E}$ and decoder $\mathcal{D}$, a U-Net $\epsilon_{\theta}$ with trainable parameter $\theta$, and a CLIP encoder $\mathcal{T}$. During training, the encoder $\mathcal{E}$ maps a training image $I \in \mathbb{R}^{H\times W \times 3}$ from pixel space to the latent space in $\mathbb{R}^{h\times w \times 4}$, where $h = \frac{H}{8}$ and $w = \frac{W}{8}$. Then, $\epsilon_{\theta}$ is trained using the following loss function:
\begin{equation}
 \mathcal{L} = \mathbb{E}_{\mathcal{E}(I),\epsilon \sim \mathcal{N}(0,1),t} [\| \epsilon - \epsilon_{\theta}(z_{t},t,c) \|_{2}^{2}],
\end{equation}
where $t$ represents the time step, $c = \mathcal{T}(Y)$ is the condition extracted from caption $Y$, $z_t$ is the latent $\mathcal{E}(I)$ with Gaussian noise $\epsilon \sim \mathcal{N}(0,1)$ added stochastically.
During inference, a latent is sampled from Gaussian noise, which is then denoised by $\epsilon_{\theta}$ under the guidance of $c$ for $T$ time steps. Finally, the decoder $\mathcal{D}$ maps the denoised latent back to the pixel space to get the generated image.

\section{More Details about Datasets}
We conducted experiments on the DeepFashion Multimodal~\cite{liuLQWTcvpr16DeepFashion}, SHHQ~\cite{fu2022styleganhuman} and VITON-HD~\cite{choi2021viton} datasets:
\begin{itemize}
    \item DeepFashion-Multimodal contains 12,701 full-body images and their corresponding text descriptions. 
    Following FashionTex~\cite{lin2023fashiontex}, we randomly divided the images into 11,265 for the training set and 1,136 for the testing set. 
    \item SHHQ is composed of 40,000 full-body images, with the first 35,000 images used as the training set, and the remaining as the testing set. 
    \item VITON-HD comprises 11,647 images for the training set and 2,032 images for the testing set. 
\end{itemize}

\section{More Details about Feature Clustering}

As mentioned in Sec.~4.2 of the main paper, we cluster the CLIP features into eight categories to balance their contributions.
For this purpose, we compute the average cluster groups over 40,000 pieces of clothing and apply the same grouping strategy to all samples, which is not only effective in identifying representative features but also computationally efficient.

\section{More Details about User Study}

To complement Tables 2 and 4 of the main paper, we use a different method for comparison which assigns numerical scores to the rankings as follows: For $n$ methods, the 1st rank gets $n$ points, 2nd rank gets $n-1$ points, and 3rd rank gets $n-2$ points, etc. 
As shown in Table \ref{tab:user_study}, our method still receives the highest average score compared to SOTA ones, indicating that our method is the most favorable by users.




\section{Additional Qualitative Results}

\noindent\textbf{ucVTON.} Figs. \ref{fig:more1}, \ref{fig:more2}, \ref{fig:more3} and \ref{fig:more4} show additional qualitative results of our method using different style and texture conditions, which further demonstrates its robustness and generalizability. 
Moreover, Figs. \ref{fig:more1} and \ref{fig:more2} use the same style inputs, but with the texture inputs in reversed order.
This demonstrates that our method successfully disentangles the texture from the input texture images without being affected by their styles.


\vspace{1mm}
\noindent\textbf{Demo.} 
We include a demo in the supplementary materials to explain our idea in a more intuitive way. 

\vspace{1mm}
\noindent\textbf{Texture Transfer.} 
In Fig. \ref{fig:texture_comparison_visual_supp}, we show additional results on garment texture transfer, which further demonstrates that our method achieves excellent texture transfer results on a wide range of texture patterns (\eg, pure color patterns, floral patterns, stripe patterns and plaid patterns).

\vspace{1mm}
\noindent\textbf{In-shop Virtual Try-on.} 
As Fig. \ref{fig:try_on_supp} shows, our method achieves comparable results to SOTA ones. 

\section{Limitations and Future Work} 
For future work, we plan to explore adding user controls over garment shape and fit to further improve the virtual try-on experience. We hope our work will inspire more research into unconstrained VTON to enable highly customizable and personalized outcomes.

\end{document}